\documentclass[10pt,letterpaper,twocolumn,teaser,logo]{planarxiv}
\usepackage{xspace}
\usepackage{tabularx}
\newcommand{\methodname}{DexTacWAM}

\newcommand{\ie}{\textit{i.e.},\xspace}      

\definecolor{dexnavy}{HTML}{14233F}
\definecolor{tacblue}{HTML}{275CDA}

\newcommand{\dextacwam}{%
  \textbf{\textcolor{dexnavy}{Dex}%
  \textcolor{tacblue}{Tac}%
  \textcolor{dexnavy}{WAM}}}

\title{\dextacwam: A Visuo-Tactile World-Action Model for Dexterous Manipulation}
\runningtitle{\dextacwam: A Visuo-Tactile World-Action Model for Dexterous Manipulation}

\newcommand{\affmark}[1]{\textsuperscript{\textcolor{PlanAccent}{#1}}}
\authorblock{\normalfont\fontsize{10}{13}\selectfont
  Haoran Yuan\affmark{1,\ensuremath{\ddagger}}\quad
  Zekai Wang\affmark{2}\quad
  Boning Shao\affmark{2}\quad
  Haoran Lu\affmark{3}\\[3pt]
  Trevor Darrell\affmark{2}\quad
  Ismini Lourentzou\affmark{1,\ensuremath{\dagger}}\quad
  Wei Zhan\affmark{2,\ensuremath{\dagger}}\\[5pt]
  {\fontsize{9}{11}\selectfont
  \textsuperscript{1}University of Illinois Urbana-Champaign\quad
  \textsuperscript{2}University of California, Berkeley\quad
  \textsuperscript{3}Northwestern University}
}
\authornote{%
  \textsuperscript{\ensuremath{\ddagger}}Project lead\qquad
  \textsuperscript{\ensuremath{\dagger}}Equal advising, co-corresponding authors\\[2pt]
  \href{mailto:lourent2@illinois.edu}{\texttt{lourent2@illinois.edu}}\quad
  \href{mailto:wzhan@berkeley.edu}{\texttt{wzhan@berkeley.edu}}
}
\headerlogos{\institutionlogos}
\hypersetup{pdfauthor={Haoran Yuan, Zekai Wang, Boning Shao, Haoran Lu, Trevor Darrell, Ismini Lourentzou, Wei Zhan}}
\projecturl{https://dextacwam.github.io/}
\paperlinks{\projectlinks}

\teaserposition{afterabstract}
\teaser{%
\includegraphics[width=\linewidth,trim=5bp 15bp 5bp 15bp,clip]{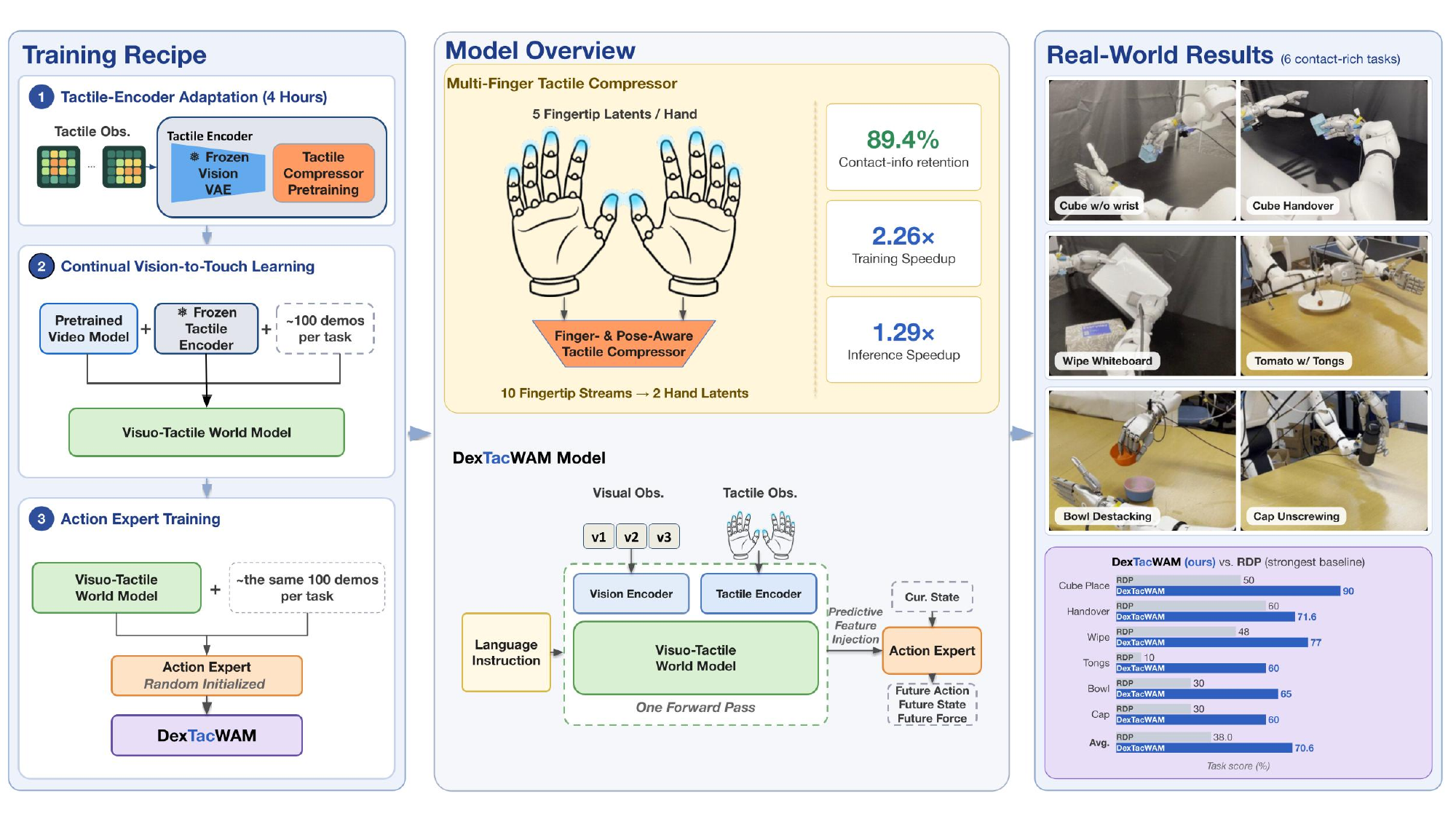}
  \vspace{-0.5cm}
  \caption{We introduce \dextacwam, a \underline{\textbf{Dex}}terous \underline{\textbf{Tac}}tile \underline{\textbf{W}}orld-\underline{\textbf{A}}ction \underline{\textbf{M}}odel that extends a pretrained video world model to multi-finger tactile dynamics through lightweight tactile adaptation and exposes predictive visuo-tactile features directly to the action expert. \methodname{} enables contact-aware control in dexterous manipulation tasks requiring occlusion reasoning, handover, and sustained multi-finger contact.\looseness-1}
  \label{fig:teaser}
}
\begin{document}
\begin{abstract}
Dexterous manipulation depends on contact dynamics that are often only partially observable from vision.
Recent World-Action Models (WAMs) couple predictive video world modeling with action generation, but remain largely vision-centric and therefore cannot directly model these contact dynamics.
We present \dextacwam{}, a visuo-tactile WAM that encodes each fingertip independently, aggregates the resulting features through a finger- and pose-aware tactile compressor, and injects the tactile latent into a video diffusion world model for joint visuo-tactile world modeling.
Across six contact-rich dexterous manipulation tasks on a 22-DoF bimanual platform, \methodname{} achieves the highest score on every task, averaging $70.6$ versus $38.0$ for the strongest baseline.
Ablations attribute the gain to modeling contact evolution as part of the predicted world state rather than tactile conditioning alone: removing tactile world modeling reduces the four-task mean from $74.7$ to $26.6$ while keeping the same tactile features and action expert.
After four hours of tactile-encoder adaptation with a frozen pretrained vision VAE, our continual vision-to-touch learning extends the pretrained video model to touch using roughly 100 demonstrations per task \emph{without tactile midtraining}, while retaining visual prediction quality within $0.5$\,dB of vision-only counterparts.
The compressor retains $89.4\%$ of pre-fusion contact recall while enabling $2.26\times$ faster training and $1.29\times$ faster inference.
Together, these results show that pretrained video priors can be extended to distributed multi-finger contact dynamics in a data- and compute-efficient manner.
\end{abstract}
\maketitlepage
\setlength{\parskip}{3pt plus 1pt minus 1pt}

\section{Introduction}
\label{sec:introduction}

World models and World-Action Models (WAMs) have emerged as a promising paradigm for robot learning, enabling agents to predict action-conditioned futures and generate actions through learned latent dynamics~\cite{ha2018worldmodels,hafner2023dreamerv3,bruce2024genie,ye2026dreamzero}. Recent robot foundation models and vision-language-action (VLA) policies further demonstrate the potential of large-scale visual representations for general robot control~\cite{brohan2022rt1,zitkovich2023rt2,ghosh2024octo,kim2024openvla,black2024pi0}. However, most of these models remain largely vision-centric, modeling how the visible scene evolves, but not how hidden physical interaction states evolve. This is a critical limitation for contact-rich manipulation, where task success depends on physical variables such as contact, slip, and grasp stability that are only partially observable from cameras~\cite{suresh2024neuralfeels,huang2024vitac3d}. For dexterous hands, the challenge is compounded by the need to coordinate contacts across multiple fingers as hand configuration and hand-object contact geometry change ~\cite{yin2025dexgen,huang2025spatiallyat}.\looseness-1

Recent works have begun to explore visuo-tactile world modeling by incorporating tactile signals into predictive robot models~\cite{yuan2026vtam,higuera2026vtwm,huang2025tactilevla,bi2025vlatouch,cheng2025omnivtla,liu2025mla}. These works demonstrate that tactile information can improve world modeling for contact-rich manipulation and validate the importance of predicting both visual and tactile observations. However, existing visuo-tactile world modeling efforts are primarily studied with parallel grippers or relatively simple end-effector setups~\cite{yuan2026vtam,higuera2026vtwm,zheng2026omnivta,huang2024vitac3d,xue2025reactive}, where tactile sensing is localized and structurally much simpler than the distributed multi-finger tactile observations of dexterous hands. As a result, it remains unclear how to extend visuo-tactile world modeling to dexterous manipulation, where tactile signals must preserve finger identity, contact locality, temporal structure, and hand-pose-dependent contact dynamics.\looseness-1

A further challenge is the asymmetry between visual and tactile data: human video is abundant, whereas tactile data remain scarce and expensive to collect. Unlike passive visual recording, tactile acquisition requires physical interaction, sensor contact, and often heterogeneous hardware~\cite{higuera2024sparsh,fu2024tvl,guzey2023dexterity,sferrazza2023power}. These constraints make large-scale tactile pretraining costly and motivate the reuse of representations and predictive priors learned from abundant visual data.

In this work, we introduce \dextacwam{}, a visuo-tactile WAM for dexterous manipulation. Our key idea is to treat tactile sensing as part of the modeled world state rather than solely using tactile observations as an auxiliary policy input. \methodname{} extends video-based WAMs to dexterous visuo-tactile co-world modeling, jointly learning visual scene dynamics and multi-finger tactile contact dynamics. To support dexterous hands, \methodname{} encodes tactile observations from each finger independently and aggregates them through a finger- and pose-aware tactile compressor to form compact hand-level tactile latents that are injected into a video diffusion world model. The resulting joint visuo-tactile world state is then used by a downstream action expert to generate dexterous actions informed by both visible scene dynamics and evolving tactile contact dynamics.

Given this asymmetry, we acquire these tactile predictive capabilities through a \textbf{continual vision-to-touch learning} strategy that extends a model pretrained on visual data to predict tactile contact dynamics while retaining its visual prediction capability. We first adapt lightweight tactile-encoder components using approximately just four hours of recorded interactions, keeping the pretrained visual VAE frozen. We then freeze the tactile encoder and fine-tune the pretrained video backbone for joint visual and tactile prediction using roughly 100 visuo-tactile teleoperation demonstrations per task. The same task demonstrations are used to train the action expert. This approach builds on pretrained visual priors and temporal video dynamics to learn tactile prediction from limited interaction data while preserving visual prediction quality.

\noindent Our contributions are summarized as follows:
\begin{itemize}[itemsep=0.5ex, parsep=0pt, topsep=-2.3pt, leftmargin=0.7cm]
\item[\textbf{(1)}] We introduce \dextacwam{}, a visuo-tactile world-action model for dexterous manipulation that jointly models visual scene dynamics and distributed multi-finger contact dynamics.
\item[\textbf{(2)}] We propose a \textbf{finger- and pose-aware tactile compressor} that preserves finger identity and localized contact information while mapping five fingertip streams into one latent per hand. This compression reduces computational cost, yielding 2.26$\times$ faster training and 1.29$\times$  faster inference.
\item[\textbf{(3)}] We introduce \textbf{continual vision-to-touch learning}, a data- and compute-efficient strategy that extends pretrained visual models with tactile predictive capabilities while maintaining visual prediction quality comparable to vision-only models.
\item[\textbf{(4)}] We evaluate on six contact-rich dexterous manipulation tasks on real hardware under a protocol that grants every method identical camera access. \methodname{} achieves the highest score on every task and a mean task score of 70.6 versus 38.0 for the strongest baseline.
Controlled ablations show that the gain comes from modeling contact evolution as part of the predicted world state rather than from tactile conditioning alone.
\end{itemize}
\section{Related Work}
\label{sec:related_work}

\subsection{World Models and WAMs}
World models learn predictive representations of environment dynamics, enabling agents to reason about future observations, latent state transitions, and action-conditioned rollouts~\cite{ha2018worldmodels,hafner2023dreamerv3,bruce2024genie,ye2026dreamzero}. In robotics, recent VLA models and generalist robot policies have demonstrated the potential of large-scale visual representations for action generation across diverse manipulation tasks~\cite{brohan2022rt1,zitkovich2023rt2,ghosh2024octo,kim2024openvla,black2024pi0}. However, these models are predominantly vision-centric, focusing on how the visible scene evolves or how visual-language features map to actions, without explicitly representing the physical interaction states that govern contact-rich control. Contact, slip, and grasp stability are often hidden or only weakly observable from cameras, but they are central to dexterous manipulation, where contact is distributed across multiple fingers and changes with hand pose and object geometry.
\methodname{} addresses this gap by extending video-based WAMs to joint visuo-tactile co-world modeling with distributed multi-finger tactile dynamics.

\subsection{Visuo-Tactile World Modeling}
Recent works have begun to incorporate tactile sensing into predictive robot models and world-action frameworks~\cite{yuan2026vtam,higuera2026vtwm,zheng2026omnivta}.
VTAM augments a pretrained video-action model with tactile streams for contact-rich manipulation, showing that tactile feedback can correct visual estimation errors and improve action generation~\cite{yuan2026vtam}. VT-WM predicts future visual and tactile observations, demonstrating that tactile grounding improves the physical fidelity of imagined rollouts under occlusion and ambiguous contact~\cite{higuera2026vtwm}. OmniVTA further explores visuo-tactile world modeling with a larger contact-rich manipulation dataset and a world-model-based framework for predicting short-horizon contact evolution~\cite{zheng2026omnivta}. 

These works validate the importance of tactile sensing for contact-rich world modeling. However, existing visuo-tactile world modeling efforts are primarily studied with parallel grippers or relatively simple end-effector setups~\cite{yuan2026vtam,higuera2026vtwm,zheng2026omnivta,huang2024vitac3d,xue2025reactive}, where tactile observations are localized and structurally simpler than those from dexterous hands. In contrast, \methodname{} targets dexterous manipulation, where tactile signals are distributed across multiple fingers, and the model must preserve finger identity, contact locality, temporal structure, and hand-pose-dependent contact dynamics.

Closest to our setting, ViTacFormer autoregressively forecasts future tactile tokens through an auxiliary head inside its ACT/CVAE policy and feeds the completed forecast back for action generation, without predicting future vision~\cite{heng2025vitacformer}. Instead, \methodname{} jointly models future visual and tactile latents as a visuo-tactile world state on which a separate action expert conditions, so touch is predicted jointly with the scene. 
Its action expert receives predictive features from a single world-model forward pass, without requiring fully denoised future visual or tactile observations.
This creates a prediction-depth versus latency trade-off absent in forecast-then-act designs.

\subsection{Tactile Dexterous Manipulation}
Tactile sensing provides direct information about local interaction forces, contact geometry, slip, and grasp stability, making it a critical modality for contact-rich manipulation~\cite{suresh2024neuralfeels,higuera2024sparsh,fu2024tvl,sferrazza2023power}. Prior tactile robot learning methods often use tactile observations as policy inputs, auxiliary perception features, or feedback signals for control~\cite{huang2024vitac3d,yu2025forcevla,guzey2023dexterity,trex2026}. While effective for many manipulation settings, this treatment does not explicitly make tactile contact dynamics part of the modeled world state. For dexterous hands, this distinction is important: tactile observations are distributed across fingers, evolve, and are tightly coupled with hand pose, object motion, and contact geometry~\cite{huang2025spatiallyat,sun2025vtaobimanip}. 
Compressing these streams without accounting for their structure can discard localized contact information, while directly mixing visual and tactile tokens with mismatched distributions can make joint action-model optimization poorly conditioned~\cite{peng2022balanced}. \methodname{} therefore introduces a structured multi-finger tactile pathway that encodes each finger independently, aggregates tactile information through a finger- and pose-aware tactile compressor, and injects the resulting tactile latent into the world model for visuo-tactile co-world modeling.\looseness-1
\begin{figure*}[t!]
    \centering
    \includegraphics[width=0.99\textwidth,trim=5bp 10bp 5bp 10bp,clip]{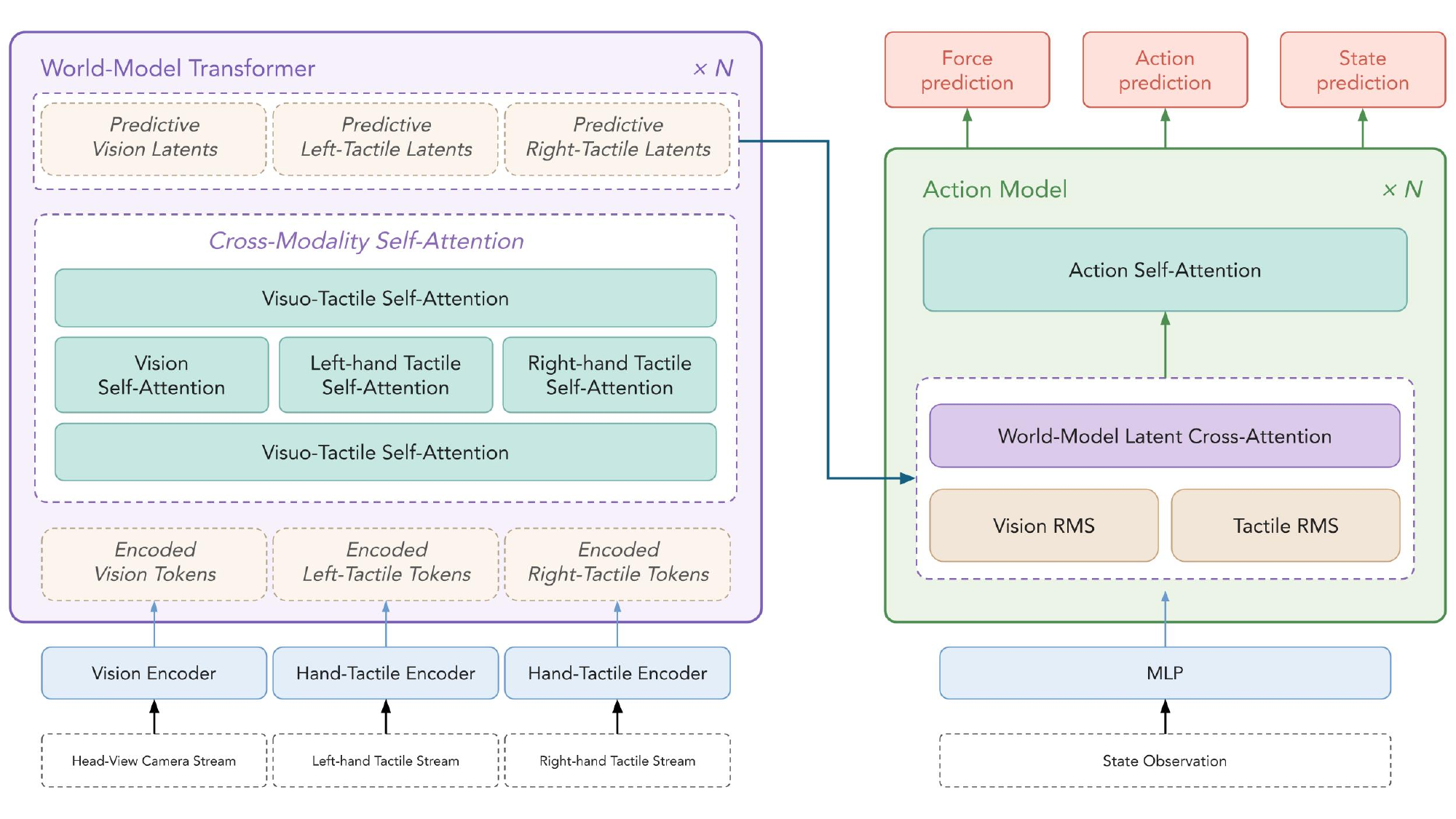}
    \vspace{-0.2cm}
    \caption{\textbf{Overall architecture of \dextacwam{}.}
    A two-block transformer design.
    \textbf{Left:} the World-Model Transformer ($\times N$) fuses vision and bimanual tactile tokens through per-modality self-attention and a shared cross-modality self-attention, producing predictive latents that form a joint visuo-tactile world state.
    \textbf{Right:} the Action Model ($\times N$) cross-attends to the World-Model latents with per-modality K/V RMS normalization, conditioned on a state token from proprioception, and predicts contact force, action, and next-state.
    Inputs are head-view RGB and left/right dexterous tactile streams encoded by lightweight encoders, together with raw state observations.}
    \label{fig:architecture}
\end{figure*}
\section{Method}
\label{sec:method}

We consider bimanual dexterous manipulation with multi-view visual observations, multi-finger tactile observations, hand proprioception, and robot actions. At each time step $t$, the robot observes RGB images $o_t^v$, tactile observations $o_t^\tau$ distributed across fingers and hands, proprioceptive hand state $q_t$, and executes an action $a_t$. 

Our goal is to learn a visuo-tactile WAM that jointly models visual scene dynamics and tactile contact dynamics, and uses the resulting joint visuo-tactile world state for contact-aware dexterous action generation.
Unlike vision-centric WAMs that only model future visual latents, \methodname{} treats tactile contacts as part of the modeled world state. Let $z_t^v$ denote visual latents and $z_t^\tau\!=\!\{z_t^{\tau,L},\hat z_t^{\tau,R}\}$ denote  compressed left- and right-hand tactile latents. We define the joint visuo-tactile world state as $s_t\!=\!\{z_t^v, z_t^\tau\}.$
As shown in Figure~\ref{fig:architecture}, the world model learns dynamics over this joint latent state, while the action model conditions on the learned visuo-tactile representation to predict dexterous actions.

\begin{figure*}[t]
    \centering
    \includegraphics[width=0.9\textwidth,trim=5bp 15bp 5bp 35bp,clip]{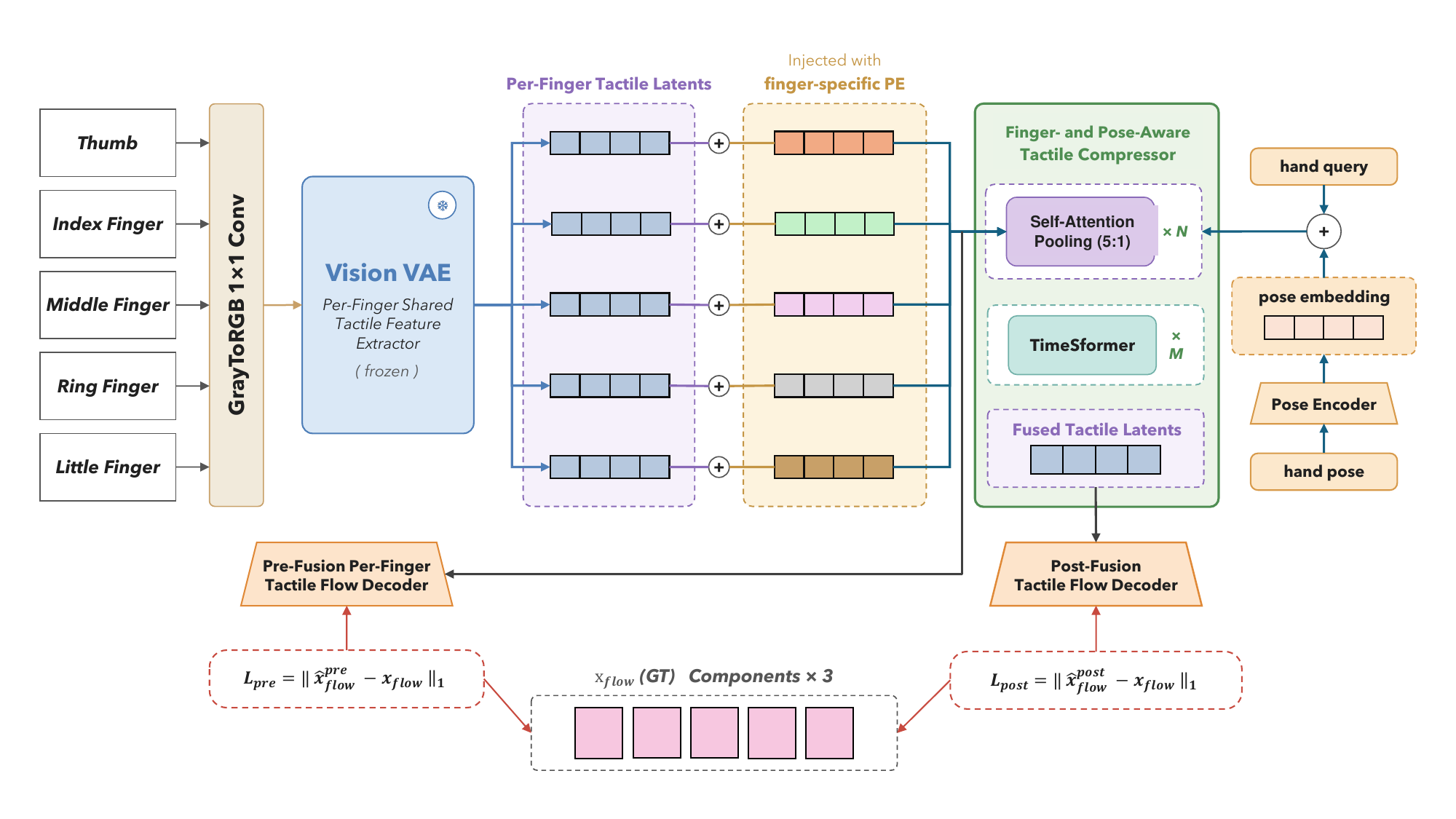}
    \vspace{-0.4cm}
    \caption{\textbf{Multi-finger tactile encoder with a finger- and pose-aware tactile compressor.}
Per-finger tactile observations are converted to pseudo-RGB inputs and encoded by a frozen pretrained vision VAE shared across fingers. Finger-specific positional embeddings preserve finger identity before $5{:}1$ self-attention pooling compresses the five tactile streams into one hand latent using a hand query and pose embeddings, followed by temporal refinement. Pre- and post-fusion reconstruction losses encourage the compressed tactile latent to retain localized multi-finger contact information.}
    \label{fig:tactile_encoder}
\end{figure*}

\subsection{Multi-Finger Tactile Encoder}
\label{sec:tactile_encoder}
A key design choice in \methodname{} is to reuse a pretrained visual VAE as the per-finger tactile feature extractor instead of training a tactile VAE from scratch. Dense tactile maps contain spatially organized local structures such as contact regions, deformation patterns, edges, and pressure-sensitive texture changes, making them compatible with image-style latent encoding. Reusing the visual VAE provides a strong spatial prior and maps tactile observations into a latent space compatible with the video diffusion world model. Appendix Figure~\ref{fig:raw_tactile_observations} shows representative raw tactile observations before encoding.

Each per-finger tactile map $o_{t,i}^{\tau}$ is single-channel, while the visual VAE expects three-channel inputs. We therefore use a lightweight $1{\times}1$ grayscale-to-RGB adapter $\phi_{g\rightarrow rgb}$ initialized to replicate the grayscale tactile signal across RGB channels. As shown in Figure \ref{fig:tactile_encoder}, the tactile map is then encoded by the frozen visual VAE encoder 
\begin{equation}
    z_{t,i}^{\tau}
    =
    E_{\mathrm{VAE}}\!\left(\phi_{g\rightarrow rgb}(o_{t,i}^{\tau})\right),
\end{equation}
where $i$ indexes fingers. For each hand, this produces a set of per-finger tactile latents $Z_t^\tau\!=\!\{z_{t,1}^{\tau}, z_{t,2}^{\tau}, \ldots, z_{t,N}^{\tau}\}.$
In the bimanual setting, tactile latents are produced for both left and right hands, preserving distributed multi-finger contact information before global aggregation.

\subsection{Finger- and Pose-Aware Tactile Compression}
\label{sec:tactile_compressor}
Processing all ten fingertip streams as separate world-model inputs increases the token count and computational cost. We therefore introduce a finger- and pose-aware tactile compressor that aggregates per-finger latents into compact hand-level tactile latents while preserving finger identity, hand pose, and temporal contact evolution.
The compressor first augments tactile latents with learnable finger identity embeddings $\tilde{z}_{t,i}^{\tau} = z_{t,i}^{\tau} + e_i,$
where $e_i$ denotes the embedding of finger $i$. 

To disambiguate contacts under different hand configurations, we inject hand pose information through a pose encoder $b_t^q\!=\!E_q(q_t),$
which is added to the tactile aggregation query or intermediate tactile tokens. At each spatio-temporal cell, self-attention pooling processes one hand query and five finger tokens, then retains the hand-query output to perform the $5{:}1$ compression. The compressor also uses lightweight divided space-time attention to model contact evolution over time, which is important for slip, regrasping, handover, and other temporally defined contact events. The compressor outputs compact bimanual tactile latents
\begin{equation}
\setlength{\abovedisplayskip}{5pt}
\setlength{\belowdisplayskip}{5pt}
\setlength{\abovedisplayshortskip}{5pt}
\setlength{\belowdisplayshortskip}{5pt}
\begin{gathered}
    \hat{z}_{t}^{\tau,L}, \hat{z}_{t}^{\tau,R}
    =
    A_{\theta}(Z_t^{\tau,L}, Z_t^{\tau,R}, q_t),
    \\
    \hat{z}_{t}^{\tau,\cdot}
    \in
    \mathbb{R}^{C \times T_{\mathrm{lat}} \times H' \times W'} ,
\end{gathered}
\end{equation}
where $L$ and $R$ denote the left and right hands. The temporal axis $T_{\mathrm{lat}}$ is aligned with the visual latent sequence, allowing each compressed hand representation to be integrated into the world model as an additional view.\looseness-1

\subsection{Visuo-Bimanual Tactile World Modeling}
\label{sec:world_model}
 \methodname{} extends video diffusion world modeling to the bimanual visuo-tactile setting. Instead of treating tactile signals as auxiliary policy inputs, we include tactile latents as part of the modeled world state $  s_t\!=\!\{z_t^v, \hat{z}_{t}^{\tau,L}, \hat{z}_{t}^{\tau,R}\}.$
Concretely, the left- and right-hand tactile latents are appended to visual latents along the view axis, so the diffusion transformer receives $V\!=\!V_v + V_\tau$ latent views, where $V_v$ are visual camera views and $V_\tau$ are tactile views. 
The DiT architecture remains unchanged, \ie cross-view self-attention couples visual and tactile tokens, enabling joint denoising of both modalities to model scene evolution and contact dynamics.\looseness-1

Let $s^{\mathrm{clean}}\!=\![z^v \Vert \hat{z}^{\tau}]$ denote the clean visuo-tactile latent state. During world-model training, we sample noise level $\sigma \in (0,1]$, corrupt the latent state with Gaussian noise $\epsilon$, and train the diffusion transformer to predict the flow-matching velocity $v_\theta(s_\sigma,\sigma) \approx \epsilon - s^{\mathrm{clean}} .$
We use a modality-split denoising objective
\begin{equation}
\setlength{\abovedisplayskip}{5pt}
\setlength{\belowdisplayskip}{5pt}
\setlength{\abovedisplayshortskip}{5pt}
\setlength{\belowdisplayshortskip}{5pt}
\begin{aligned}\label{eq:wm}
    \mathcal{L}_{\mathrm{wm}}
    &=
    \lambda_v
    \mathbb{E}
    \left[
    \left\|
    v_\theta(s_\sigma,\sigma)_{[:V_v]}
    -
    (\epsilon - z^v)
    \right\|_2^2
    \right]
    \\
    &\quad
    +
    \lambda_\tau
    \mathbb{E}
    \left[
    \left\|
    v_\theta(s_\sigma,\sigma)_{[V_v:]}
    -
    (\epsilon - \hat{z}^{\tau})
    \right\|_2^2
    \right],
\end{aligned}
\end{equation}
that encourages the world model to preserve both global visual scene evolution and local tactile contact evolution.

\subsection{Per-Modality K/V Normalized Action Expert}
\label{sec:action_model}
On top of the visuo-tactile world model, \methodname{} trains an action expert for dexterous action generation. A central challenge is that visual and tactile latents have substantially different token distributions and scales: visual latents are dense and carry global scene information, while tactile latents are sparse and contact-sensitive. Naively concatenating them in a shared cross-attention space results in inconsistent K/V distributions across modalities and makes action-model optimization unstable; in our experiments, the action model fails to converge without modality-wise normalization.

To address this, we use a shared action cross-attention with per-modality K/V RMS normalization. Before action cross-attention, we split the K/V sequence into visual and tactile partitions, normalize each partition independently, and concatenate them back:
\begin{equation}
\setlength{\abovedisplayskip}{5pt}
\setlength{\belowdisplayskip}{5pt}
\setlength{\abovedisplayshortskip}{5pt}
\setlength{\belowdisplayshortskip}{5pt}
\begin{gathered}
    \widetilde{x}
    =
    \Big[
    \mathrm{RMS}_v(x_{[:V_vL]})
    \;\Vert\;
    \mathrm{RMS}_\tau(x_{[V_vL:]})
    \Big],
    \\
    h^a = \mathrm{Attn}^a(h,\widetilde{x}) .
\end{gathered}
\end{equation}
The attention module itself is shared across modalities, and only the pre-attention K/V statistics are normalized separately. We use parameter-free RMSNorm~\cite{zhang2019root}, so this design introduces no learnable modality gates or separate visual/tactile action pathways. This keeps the action expert simple while reducing visual dominance in the attention softmax.

The action target is a continuous vector containing per-finger force, arm pose, hand target, and proprioceptive state prediction $    a_t =
    [
    f_t^{1:N},
    a_t^{\mathrm{arm}},
    a_t^{\mathrm{hand}},
    s_t^{\mathrm{prop}}
    ]$
where the force block contains 60 dimensions corresponding to $N=10$ fingers, each with a 6D force-related target. The arm block $a_t^{\mathrm{arm}}$ is parameterized as an end-effector pose relative to the measured pose at the start of the chunk, rather than as an absolute joint target. This keeps the prediction target centered and scale-stable across the workspace, which matters most on tasks requiring fine relative alignment such as bimanual handover. We train the action expert with a single flow-matching MSE over the full action vector:
\begin{equation}\label{eq:act}
    \mathcal{L}_{\mathrm{act}}
    =
    \mathbb{E}_{\sigma}
    \left[
    \left\|
    v_\theta^a(h,\sigma) - (\epsilon_a - a_t)
    \right\|_2^2
    \right].
\end{equation}
Thus, per-finger force prediction is directly supervised as part of the unified action target, rather than through an additional separately weighted force loss.

\begin{table*}[t!]
\centering
\caption{
\textbf{Task scores (\%) on six contact-rich dexterous manipulation tasks}, over 20 real-robot trials per method per task.
H/L/R denote head, left-wrist, and right-wrist cameras; camera access is identical across methods within each task.
Baselines use their original action parameterization. \methodname{} uses relative end-effector actions. Best results are highlighted in bold.
}
\label{tab:success_rate}
\vspace{-0.2cm}
\setlength{\tabcolsep}{10pt}
\renewcommand{\arraystretch}{1.12}
\resizebox{\linewidth}{!}{
\begin{tabular}{lccccccc}
\toprule
Method
& \begin{tabular}[c]{@{}c@{}}Cube Place\\(H)\end{tabular}
& \begin{tabular}[c]{@{}c@{}}Handover\\(H+L+R)\end{tabular}
& \begin{tabular}[c]{@{}c@{}}Wipe\\(H+L+R)\end{tabular}
& \begin{tabular}[c]{@{}c@{}}Tongs\\(H+R)\end{tabular}
& \begin{tabular}[c]{@{}c@{}}Bowl\\(H+R)\end{tabular}
& \begin{tabular}[c]{@{}c@{}}Bottle Cap\\(H+L+R)\end{tabular}
& \textbf{\textcolor{dexnavy}{Avg.}} \\
\midrule
$\pi_{0.5}$ (no tactile)
& 10.0 & 46.0 & 24.0 & 0.0 & 40.0 & 45.0 & 27.5 \\
ViTacFormer
& 20.0 & 33.3 & 26.0 & 0.0 & 20.0 & 10.0 & 18.2 \\
RDP
& 50.0 & 60.0 & 48.0 & 10.0 & 30.0 & 30.0 & 38.0 \\
Genie Envisioner (no tactile)
& 10.0 & 30.0 & 44.0 & 5.0 & 35.0 & 40.0 & 27.3 \\
\rowcolor[HTML]{EDF3FD} \dextacwam{} (ours)
& \textbf{90.0} & \textbf{71.6} & \textbf{77.0} & \textbf{60.0} & \textbf{65.0} & \textbf{60.0} & \textbf{70.6} \\
\bottomrule
\end{tabular}
}
\end{table*}

\subsection{Continual Vision-to-Touch Learning}
\label{sec:data_recipe}
We train DexTacWAM in three stages that progressively extend pretrained visual representations and video dynamics to tactile prediction and dexterous control.
First, we adapt the trainable components of the tactile encoder, \ie the grayscale-to-RGB projection and tactile compressor, while keeping the visual VAE frozen. Pretraining uses a diverse-488 corpus, which contains roughly four hours and 488 episodes of bimanual dexterous manipulation. We train on episodes 0--463 and hold out episodes 464--487 for the tactile representation analyses reported in Section~\ref{sec:ablations}.
Second, the video model receives no tactile midtraining: for each downstream task, we freeze the Stage-1 tactile modules and directly finetune the pretrained video model for cross-modal world modeling on approximately 100 task demonstrations using Eq.~(\ref{eq:wm}). Third, we initialize a fresh action expert and train it on the same task demonstrations with the action flow-matching objective in Eq.~(\ref{eq:act}), without a separate action pretraining stage. This staged procedure uses the diverse interaction corpus to learn a reusable tactile representation, followed by task-specific adaptation of the world model and action expert. 

\section{Experiments}
\label{sec:experiments}

Our experiments are designed to answer four questions:
\textbf{(i)} does predictive visuo-tactile world-action modeling improve contact-rich dexterous manipulation over strong action policies with direct tactile injection and vision-only world-model baselines;
\textbf{(ii)} which components contribute to the performance gain, does the tactile compressor preserve localized multi-finger contact information, and does the world model predict contact evolution rather than simply repeating the latest tactile observation;
\textbf{(iii)} can continual vision-to-touch learning acquire tactile predictive capability from task-scale data while retaining visual prediction quality;
and \textbf{(iv)} what training-speed and end-to-end deployment-latency improvements result from tactile compression.

\noindent \textbf{Dexterous tactile sensing platform.}
We use a bimanual dexterous manipulation platform equipped with multi-finger 22-DoF Sharpa tactile sensors that provide dense contact observations over the finger surfaces, allowing the model to observe local contact activation, pressure changes, contact shifts, and contact release during manipulation.\looseness-1

\noindent \textbf{Tasks.}
We evaluate \methodname{} on six contact-rich dexterous manipulation tasks that probe complementary aspects of contact-aware control. An illustrative overview of all tasks can be found in Figure \ref{fig:teaser}.
\textbf{Cube Place w/ Occlusion} tests whether tactile sensing helps when the cube is occluded by the arm itself, whereas
\textbf{Cube Handover} requires coordinated bimanual transfer, where grasp stability is hard to infer from cameras alone,
and \textbf{Two-Hand Wipe} requires sustained contact and stable pressure over a long horizon.
\textbf{Tongs} requires precise force control for tool-mediated cherry-tomato transfer,
\textbf{Bowl} requires thumb--index friction to separate stacked bowls,
and \textbf{Bottle Cap} requires thumb--index friction plus torque for unscrewing, \ie rotational contact.
Together these tasks cover occlusion, hand-to-hand transfer, bimanual contact maintenance, tool-mediated force control, friction-dependent separation, and rotational contact.
Per-task grading rubrics are given in Appendix~\ref{appendix:rubrics}.\looseness-1

\noindent \textbf{Observation protocol.}
Camera access is set per task and is identical across all methods within a task.
Cube Place uses the head camera only, so that occlusion cannot be resolved by a close-range view, while
Tongs and Bowl use head plus right wrist, and Handover, Wipe, and Bottle Cap use head plus both wrists.

\noindent \textbf{Training data.}
We first pretrain the tactile encoder's lightweight projection and compressor on the diverse bimanual corpus while keeping its visual VAE frozen, and then freeze the entire tactile encoder. For each downstream task, we use roughly 100 demonstrations to directly finetune the pretrained video model for cross-modal world modeling and train a randomly initialized action expert. All methods use the same per-task data and their full original training schedules.

\noindent \textbf{Baselines.}
We compare against two direct policy-learning baselines and one vision-only WAM.
$\pi_{0.5}$~\cite{intelligence2025pi_} is a generalist robot policy without tactile input, representing large-scale robot policy pretraining.
ViTacFormer~\cite{heng2025vitacformer} is a strong tactile-vision policy that consumes fingertip force/torque alongside vision; it is an ACT/CVAE policy with an auxiliary tactile-forecasting head.
Reactive Diffusion Policy (RDP)~\cite{xue2025reactive} injects the ten fingertip 6-D wrenches directly into a diffusion policy without learning predictive visuo-tactile dynamics.
Genie Envisioner (GE)~\cite{liao2025genie} is a vision-only WAM, isolating what visual world modeling alone achieves when contact state is only partially visible from cameras.
Implementation details for all baselines are given in Appendix~\ref{app:strong_baseline_details}.

\subsection{Main Results on Dexterous Manipulation}
\label{sec:main_results}

\begin{figure*}[t!]
    \centering
    \includegraphics[width=0.99\linewidth]{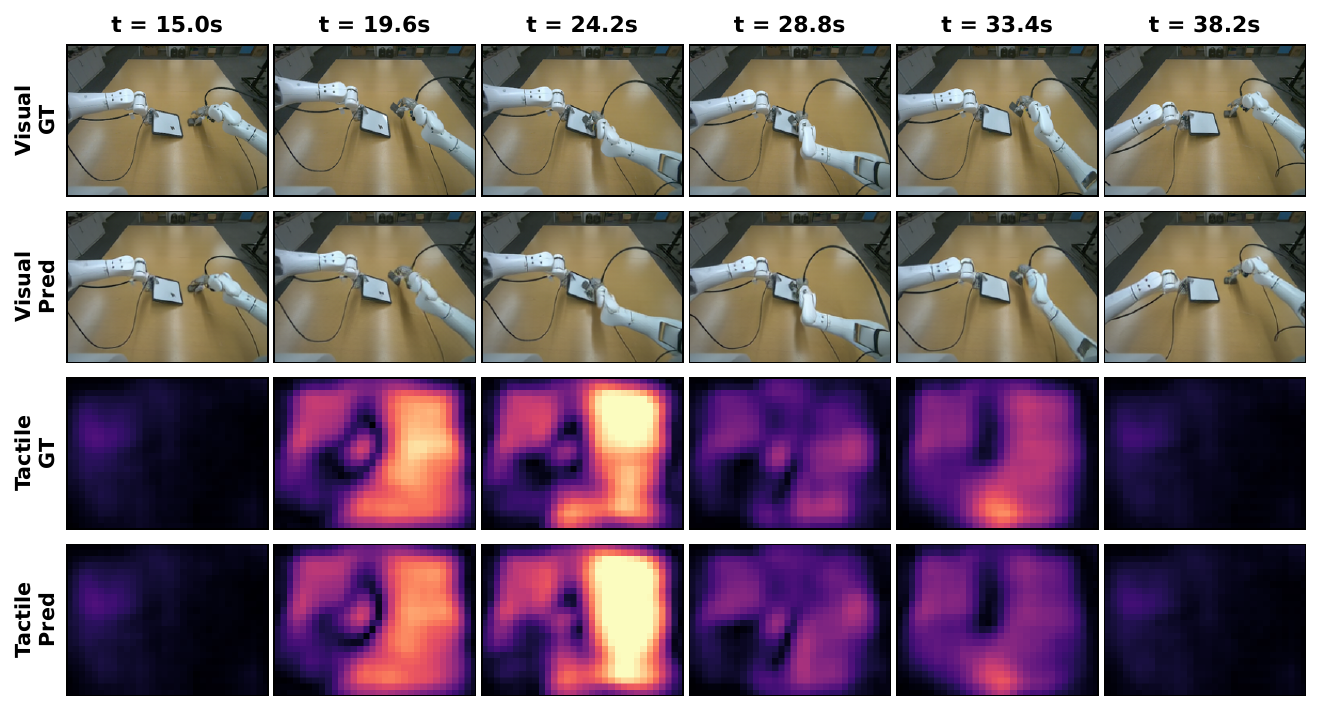}
    \vspace{-0.3cm}
    \caption{\textbf{Joint visuo-tactile world-model prediction on the wiping task.}
    Ground-truth and predicted visual and tactile observations are shown at six instants from one held-out rollout, spanning approach, contact, wiping, and release.
    The tactile rows visualize the shear magnitude of the right index fingertip using a shared color scale.
    Across consecutive prediction windows conditioned on observed history, the predicted contact patch emerges, moves, and disappears consistently with the visual interaction, rather than persisting as a static imprint.}
    \label{fig:wm_pred_visual_tactile}
\end{figure*}
Table~\ref{tab:success_rate} reports task scores over 20 real-robot trials per method per task.
\methodname{} obtains the highest score on all six tasks, with an average of $70.6$ versus $38.0$ for the strongest baseline (RDP).
These results are obtained with no tactile midtraining of the video model, an action expert trained from scratch, only four hours of tactile-encoder adaptation using a frozen pretrained vision VAE, roughly 100 demonstrations per task, and no reinforcement-learning post-training, showing that predictive tactile world modeling is effective in a limited-data regime.
The largest margin appears on Tongs ($60$ vs.\ $10$), where the task is completed through a tool, and the only reliable evidence of grasp state is the force transmitted through the tongs; every camera-driven baseline scores at or below $10$.
The pattern across tasks is consistent with the hypothesis motivating \methodname{}: the largest gains appear where task-relevant physical state is difficult to infer from instantaneous visual observations.

Comparing the two vision-only entries with the two tactile entries also suggests that tactile conditioning alone is insufficient. RDP receives the same fingertip wrenches that \methodname{} does, but consumes them as a downstream policy input rather than modeling contact evolution as part of the predicted world state, and reaches $38.0$ on average. These results suggest that jointly modeling scene and contact dynamics can provide useful predictive features for contact-rich dexterous control.

\noindent \textbf{Statistical robustness.}
To complement the per-task evaluations, we pool outcomes across the four binary tasks (Cube Place, Tongs, Bowl, and Bottle Cap), yielding $80$ trials per method. \methodname{} succeeds on $55/80$ trials ($68.8\%$), compared with $24/80$ ($30.0\%$) for the strongest baseline, RDP. The corresponding 95\% Wilson confidence intervals are $[57.9,77.8]$ and $[21.1,40.8]$, respectively, and a Fisher exact test gives $p=1.59\times10^{-6}$, indicating that \methodname{} achieves a statistically significant improvement in overall success rate over RDP.

\noindent \textbf{Generalization to unseen object configurations.}
We evaluate \methodname{} on Bowl configurations held out from training. Training uses three-bowl stacks in pink, blue, and orange. \methodname{} succeeds in $13/20$ trials on unseen green, purple, and yellow bowls, matching its in-distribution performance, and $12/20$ trials on unseen stack heights of two or four bowls. These results demonstrate robustness to changes in appearance and stack geometry. Representative rollouts are shown in Appendix~\ref{app:generalization_rollouts}.\looseness-1

\noindent \textbf{Qualitative world-model prediction.}
Figure~\ref{fig:wm_pred_visual_tactile} compares predicted and ground-truth visual and tactile futures on the wiping task.
The predicted tactile deformation evolves with the interaction: contact emerges, shifts during wiping, and fades upon release, in synchrony with the visual motion rather than remaining a static copy of the latest tactile observation.
Appendix~\ref{app:wm_prediction_visualization} expands this analysis to all fingers and tactile-flow channels, while Appendix~\ref{app:tactile_latent_tsne} examines the structure preserved by the tactile representation and compressor.

\noindent \textbf{Training and Deployment Efficiency.}
\label{sec:efficiency}
Compressing ten fingertip streams into two per-hand latents makes tactile world modeling more efficient in both training and deployment. On a single RTX~4090, it reduces the views attended by the DiT from 11 (head $+$ 10 fingertips) to 3 (head $+$ 2 hands), cutting view tokens from $3168$ to $864$ ($3.67\times$) and end-to-end latency from $363.0$ to $281.6$\,ms per chunk ($1.29\times$), below the $1.8$\,s action-chunk execution time. Under the same training setup, compression reduces iteration time from $3.87$ to $1.71$\,s/iter, a $2.26\times$ speedup that substantially lowers the cost of scaling visuo-tactile world-model training.

\begin{table}[t]
\centering
\caption{\textbf{Policy-level ablations.}
Best task score (\%) in bold.}
\label{tab:policy_ablations}
\vspace{-0.2cm}
\footnotesize
\setlength{\tabcolsep}{2pt}
\renewcommand{\arraystretch}{1.15}

\begin{tabularx}{\linewidth}{
    @{}>{\raggedright\arraybackslash}X ccccc@{}
}
\toprule
Variant
& \shortstack{Cube Place}
& Handover
& Wipe
& Tongs
& \textbf{\textcolor{dexnavy}{Avg.}} \\
\midrule

\rowcolor[HTML]{EDF3FD}
\dextacwam{}{} (full)
& \textbf{90.0}
& \textbf{71.6}
& \textbf{77.0}
& \textbf{60.0}
& \textbf{74.7} \\

w/o tactile world modeling
& 20.0
& 33.3
& 48.0
& 5.0
& 26.6 \\

\midrule
w/o per-modality K/V RMS normalization
& \multicolumn{5}{c}{\textit{Did not converge}} \\

\bottomrule
\end{tabularx}
\end{table}

\begin{table}[t!]
\centering
\caption{\textbf{Tactile-encoder component ablations.}
Higher values are better; best results are bold.}
\label{tab:adapter_recall}
\vspace{-0.2cm}
\footnotesize
\setlength{\tabcolsep}{4pt}
\renewcommand{\arraystretch}{1.15}

\begin{tabularx}{\linewidth}{
    @{}>{\raggedright\arraybackslash}X ccc@{}
}
\toprule
Variant
& \multicolumn{2}{c}{Recall $\uparrow$}
& Retention $\uparrow$ \\
\cmidrule(lr){2-3}\cmidrule(l){4-4}
& @1 & @9 & @9 \\
\midrule
Cross-attention pooling
& 0.561 & 0.492 & 0.629 \\

Self-attention pooling
& 0.656 & 0.575 & 0.717 \\

Self-attention $+$ pose injection
& \textbf{0.750} & 0.645 & 0.796 \\

Self-attention $+$ temporal refinement
& 0.748 & 0.710 & 0.873 \\

\midrule
\rowcolor[HTML]{EDF3FD}
\textbf{Full compressor} (pose $+$ temporal)
& 0.748 & \textbf{0.725} & \textbf{0.894} \\

\quad w/o finger-identity embeddings
& 0.702 & 0.671 & 0.849 \\
\bottomrule
\end{tabularx}
\end{table}

\subsection{Ablation Studies}
\label{sec:ablations}
We ablate policy-level components using task scores (Table~\ref{tab:policy_ablations}), while tactile-encoder components are evaluated by contact recall on held-out episodes (Table~\ref{tab:adapter_recall}). Contact recall provides a more direct measure of how well each encoder variant preserves contact information than downstream task scores. Implementation details for every variant are given in Appendix~\ref{app:ablation_impl}.

\noindent \textbf{Effect of tactile world modeling.}
We remove tactile prediction from the world model and instead condition the action expert directly on encoded tactile features, keeping the tactile encoder, observations, action-expert architecture, action space, and training data fixed. Across the four tasks evaluated in this ablation, the average task score decreases from $74.7\%$ to $26.6\%$ (Table~\ref{tab:success_rate}). These results show that the gain depends strongly on modeling contact evolution as part of the predicted world state rather than on tactile conditioning alone.

\noindent \textbf{Per-modality K/V RMS normalization.}
Without per-modality K/V RMS normalization in the action cross-attention, the action expert fails to converge in our experiments: open-loop predictions fail to track ground-truth actions. Differences in visual and tactile feature magnitudes can affect both the attention weights and the scale of the attention output. Per-modality K/V RMS normalization controls these magnitudes when the action expert combines the two modalities, supporting stable training.\looseness-1

\noindent \textbf{Tactile-encoder components.}
Table~\ref{tab:adapter_recall} isolates the tactile compressor components by contact recall on the 24 held-out episodes of the diverse-488 pretraining corpus.
We also measure contact recall before and after tactile fusion and define the retention ratio as
\begin{equation}
\label{eq:retention}
    \mathrm{Retention}
    =
    \frac{
    \mathrm{Post\text{-}Fusion\ Recall}
    }{
    \mathrm{Pre\text{-}Fusion\ Recall}
    }.
\end{equation}
Higher values indicate better contact information retention after compression.
Replacing hand-query cross-attention with self-attention pooling over the hand query and five finger tokens raises Recall@9 from $0.492$ to $0.575$. Adding pose injection or temporal refinement raises it to $0.645$ or $0.710$, respectively; combining both gives $0.725$ and raises retention from $62.9\%$ to $89.4\%$. Removing finger-identity embeddings from the full compressor drops Recall@9 to $0.671$, confirming that the tokens must stay finger-attributable rather than becoming an unordered contact bag. Since every variant shares the same frozen visual VAE encoder and the same per-finger inputs, these differences come from the aggregation design rather than from upstream tactile encoding. Implementation details can be found in Appendix~\ref{app:tactile_encoder_impl}.

\subsection{Analysis of Tactile Representation and World-Model Prediction}
\label{sec:quant_analysis}

We evaluate tactile prediction and whether vision-to-touch adaptation preserves visual prediction quality.\looseness-1

\noindent \textbf{Tactile world-model prediction quality.}
Table~\ref{tab:tactile_prediction} reports prediction quality on held-out trajectories. The predicted tactile futures reach $0.967$ cosine similarity and $0.063$ NMSE against ground truth, with contact precision/recall/F1 of $0.749/0.725/0.737$ in flow space. Because tactile signals are temporally smooth, a forecaster can score well by simply repeating the last observation. We therefore compare against a copy-last-frame baseline and find \methodname{} attains $0.561\times$ its error, so the model is predicting contact evolution rather than exploiting smoothness.

\begin{table}[t]
\centering
\caption{\textbf{Tactile world-model prediction quality}
on held-out trajectories. Latent metrics compare predictions with
ground-truth tactile latents; contact metrics are evaluated in flow
space. The final row reports the error ratio relative to
copy-last-frame prediction, where values below $1$ indicate improvement.}
\label{tab:tactile_prediction}
\vspace{-0.2cm}
\footnotesize
\setlength{\tabcolsep}{4pt}
\renewcommand{\arraystretch}{1.15}

\begin{tabularx}{\linewidth}{
    @{}>{\raggedright\arraybackslash}X c@{}
}
\toprule
Metric & \dextacwam{} \\
\midrule
Latent cosine similarity $\uparrow$ & 0.967 \\
Latent NMSE $\downarrow$             & 0.063 \\

\midrule
Contact precision $\uparrow$ & 0.749 \\
Contact recall $\uparrow$    & 0.725 \\
Contact F1 $\uparrow$        & 0.737 \\

\midrule
\rowcolor[HTML]{EDF3FD}
Error ratio vs.\ copy-last-frame $\downarrow$
& $\mathbf{0.561}\times$ \\
\bottomrule
\end{tabularx}
\end{table}

\noindent \textbf{Acquiring touch preserves visual prediction quality.}
\methodname{} adapts a pretrained visual world model to touch using task-scale demonstrations. We compare it against a vision-only counterpart with identical training and held-out episodes, cameras, and finetuning steps, reporting paired per-episode differences with the three views averaged within each episode. Table~\ref{tab:vision_retention} shows no detectable change on either task. The paired $95\%$ confidence intervals on $\Delta$PSNR are $[-0.16,+0.28]$\,dB on Cube Handover and $[-0.49,+0.36]$\,dB on Two-Hand Wipe, so the data bound any PSNR loss at half a decibel. Differences in SSIM and LPIPS are similarly negligible. Tactile prediction is therefore acquired while retaining visual prediction quality. Appendix~\ref{app:vision_retention_protocol} provides the full statistical protocol.\looseness-1

\section{Conclusion}
\label{sec:conclusion}
We introduce \dextacwam{}, a visuo-tactile WAM that extends pretrained video priors to jointly model visual scene dynamics and distributed multi-finger contact dynamics.  \methodname{} achieves the highest task score across all six evaluated tasks, with ablations demonstrating the benefit of explicitly modeling contact evolution beyond direct tactile conditioning. Continual vision-to-touch learning adds tactile predictive capabilities using approximately four hours of recorded interactions for tactile-encoder adaptation and roughly 100 demonstrations per task, while maintaining visual prediction quality comparable to vision-only models. The tactile compressor enables $2.26\times$ faster training and $1.29\times$ faster inference, while the action expert uses predictive features from a single world-model forward pass without requiring fully denoised future observations. These results demonstrate a data- and compute-efficient approach to extending pretrained video priors to distributed contact dynamics, providing a practical path toward WAMs for complex physical interaction.

\section{Limitations}
While \methodname{} demonstrates the value of visuo-tactile co-world modeling for dexterous manipulation, it also reveals two limitations. First, our evaluation is restricted to the Sharpa dexterous hand with vision-based tactile sensing, and transfer to other sensing principles, such as capacitive, piezoresistive, piezoelectric, or magnetic sensing, remains untested. Second, both the world model and action expert are trained on successful demonstrations per task. This success-only distribution supports strong nominal execution but provides little supervision for failure recovery or self-correction. Future work should explore broader sensor and embodiment pretraining, as well as failure/recovery data collected through perturbation, intervention, or reinforcement learning; Appendix~\ref{app:success_bias} provides further discussion.

\begin{table}[t!]
\centering
\caption{\textbf{Effect of tactile modeling on visual prediction.}
Future-RGB prediction quality for the visuo-tactile world model and its matched vision-only counterpart, with PSNR in dB.
$\Delta$ denotes visuo-tactile minus vision-only.}
\label{tab:vision_retention}
\vspace{-0.2cm}
\footnotesize
\setlength{\tabcolsep}{4pt}
\renewcommand{\arraystretch}{1.15}

\begin{tabularx}{\linewidth}{
    @{}>{\raggedright\arraybackslash}X cc@{}
}
\toprule
Metric
& \shortstack{Cube Handover\\(25K steps)}
& \shortstack{Two-Hand Wipe\\(30K steps)} \\
\midrule

Vision-only PSNR (dB) $\uparrow$
& 17.78 & 17.14 \\

Visuo-tactile PSNR (dB) $\uparrow$
& 17.84 & 17.07 \\

\addlinespace[3pt]
\rowcolor[HTML]{EDF3FD}
$\Delta$PSNR (dB) $\uparrow$
& $+0.06$ & $-0.07$ \\

\midrule
Vision-only SSIM $\uparrow$
& 0.7286 & 0.7342 \\

Visuo-tactile SSIM $\uparrow$
& 0.7294 & 0.7280 \\

\addlinespace[3pt]
\rowcolor[HTML]{EDF3FD}
$\Delta$SSIM $\uparrow$
& $+0.0008$ & $-0.0062$ \\

\midrule
Vision-only LPIPS $\downarrow$
& 0.2071 & 0.1938 \\

Visuo-tactile LPIPS $\downarrow$
& 0.2113 & 0.2033 \\

\addlinespace[3pt]
\rowcolor[HTML]{EDF3FD}
$\Delta$LPIPS $\downarrow$
& $+0.0042$ & $+0.0095$ \\

\bottomrule
\end{tabularx}
\end{table}

\begingroup
\small
\phantomsection
\addcontentsline{toc}{section}{References}
\bibliographystyle{plainnat}
\bibliography{references}
\endgroup
\newpage
\appendix
\clearpage
\newpage
\section{Appendix}
\label{app:appendix}


\subsection{Task Rollouts and Evaluation Rubrics}
\label{appendix:rubrics}
\begin{figure*}[p]
    \centering
    \includegraphics[width=\linewidth,trim=46bp 0 46bp 0,clip]{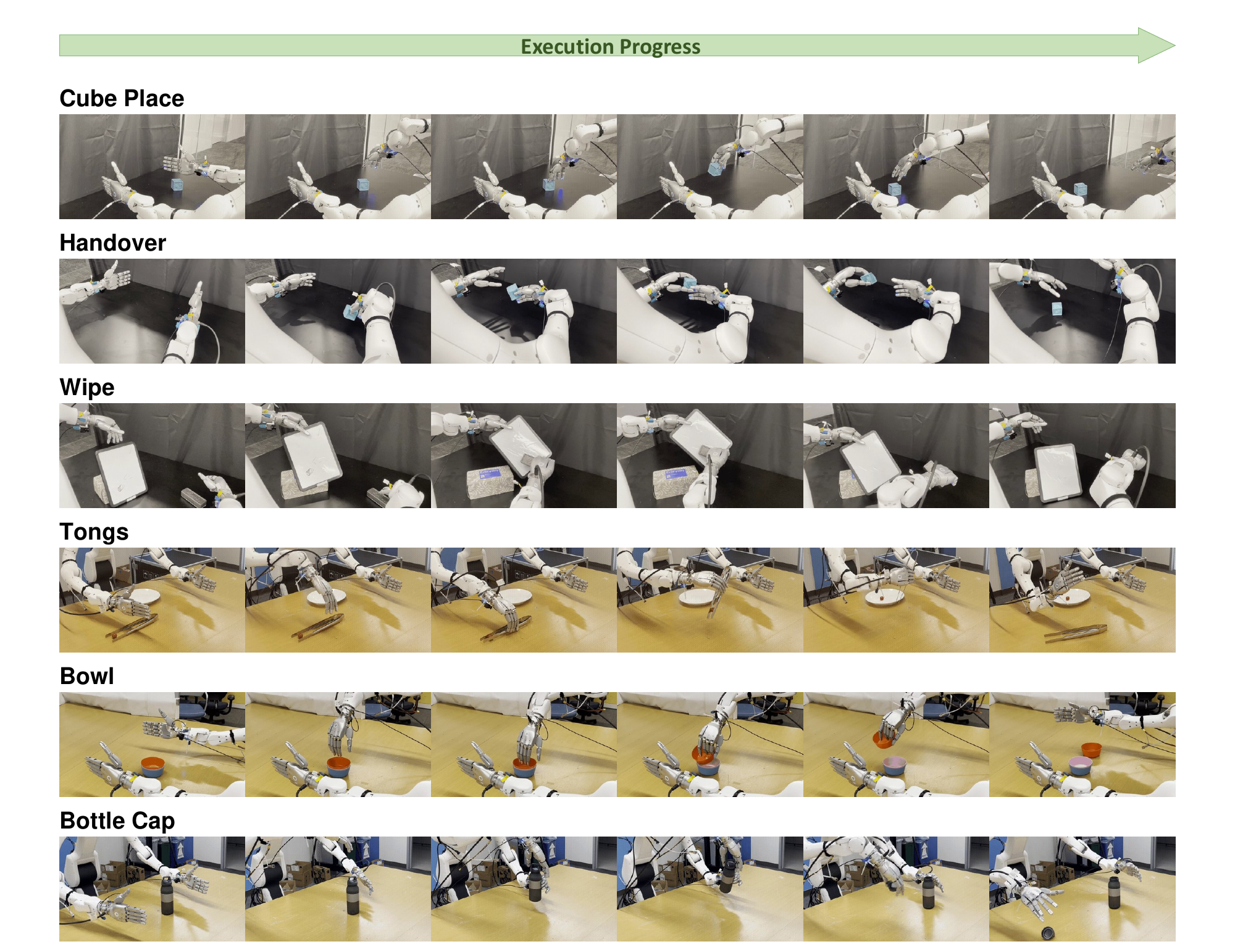}
    \vspace{-0.6cm}
    \caption{\textbf{Representative \methodname{} rollouts on the six evaluation tasks.}
    Keyframes progress from left to right and are sampled from autonomous closed-loop executions.
    The tasks span visual occlusion, bimanual transfer, sustained surface contact, tool-mediated manipulation, multi-finger separation, and rotational contact under torque.}
    \label{fig:six_task_demonstrations}
\end{figure*}

Figure~\ref{fig:six_task_demonstrations} illustrates the complete interaction sequence for each of the six real-world tasks. Each row progresses from the initial configuration on the left to task completion on the right.

\noindent\textbf{Task I: Cube Place w/ Occlusion} \textit{Text Instruction:}
``pick up the cyan cube with your right hand'' The
robot must grasp a cyan cube on the table with only proprioception, a head-camera RGB stream, and optionally fingertip tactile information. The wrist cameras are disabled, and the robot arm occludes the cube from the head camera.

\textit{Grading rubric (binary):}
\begin{itemize}
    \item $+1$: \textbf{(a)} Successfully picks up the cube from the table into the air.
\end{itemize}

\noindent\textbf{Task II: Cube Handover} \textit{Text Instruction:}
``pick up the cyan cube on the table with your left hand, hand it over to the right hand, and place the cube down on the table with your right hand'' The
robot must grasp a cyan cube on the table with its left hand, lift it up and hand it over to its right hand, and finally place the cube down onto the table without dropping the cube.

\textit{Grading rubric (additive):}
\begin{itemize}
    \item $+0.33$: \textbf{(a)} Successfully picks up the cube from the table into the air.
    \item $+0.33$: \textbf{(b)} Successfully hand the cube from the left hand over to the right hand.
    \item $+0.33$: \textbf{(c)} Place the cube down on the table with the right hand.
\end{itemize}

\noindent\textbf{Task III: Two-Hand Wipe} \textit{Text Instruction:}
``hold the white board with your left hand, pick up the black whiteboard eraser with your right hand, erase the black painting on the white board with your right hand, and finally put both objects down'' A small whiteboard is placed on the table with one side tilting up, and a whiteboard eraser is placed beside the whiteboard on the table. The robot must first use its left hand to hold the whiteboard on the side that is tilted up, grasp the whiteboard cleaner with its right hand, use the whiteboard cleaner to wipe clean the black markings on the whiteboard, and place the two objects down onto the table.

\textit{Grading rubric (additive):}
\begin{itemize}
    \item $+0.2$: \textbf{(a)} Holds the side of the whiteboard tight with the left hand
    \item $+0.2$: \textbf{(b)} Successfully grasp the whiteboard cleaner with the right hand
    \item $+0.2$: \textbf{(c)} Use the whiteboard cleaner to wipe the black markings on the whiteboard.
    \item $+0.2$: \textbf{(d)} Fully cleaned all black markings on the whiteboard.
    \item $+0.2$: \textbf{(e)} Successfully put both objects onto the table.
\end{itemize}

\noindent\textbf{Task IV: Tongs} \textit{Text Instruction:}
``pick up the tongs with your right hand, use them to transfer the cherry tomato to the plate, and return the tongs to the table'' A pair of tongs, a cherry tomato, and a plate are placed on the table. The robot must first grasp the tongs stably, close them on the tomato with enough force to prevent slipping but not so much that the tomato is crushed, transfer and release the tomato onto the plate, and finally return the tongs to the table. Because the object is manipulated through a tool, the grasp state is not directly visible and must be inferred from the force transmitted through the tongs.\looseness-1

\textit{Grading rubric (binary):}
\begin{itemize}
    \item $+1$: Completes the full sequence: grasps the tongs, transfers the intact cherry tomato onto the plate without crushing or dropping it, and returns the tongs to the table.
\end{itemize}

\noindent\textbf{Task V: Bowl} \textit{Text Instruction:}
``separate the top bowl from the stack with your right hand'' A stack of bowls is placed on the table. The robot must separate the topmost bowl from the stack and lift it away without toppling the remaining stack. Separation depends on establishing sufficient thumb--index friction against the bowl rim, which is difficult to verify visually.

\textit{Grading rubric (binary):}
\begin{itemize}
    \item $+1$: Separates the top bowl from the stack and lifts it clear without knocking over the remaining bowls.
\end{itemize}

\noindent\textbf{Task VI: Bottle Cap} \textit{Text Instruction:}
``hold the bottle with your left hand and unscrew the cap with your right hand'' A capped bottle is placed on the table. The robot must stabilize the bottle with its left hand and unscrew the cap with its right hand. The task requires both thumb--index friction and sustained torque under rotational contact, where slip is only observable through touch.

\textit{Grading rubric (binary):}
\begin{itemize}
    \item $+1$: Fully unscrews and removes the cap from the bottle while the bottle remains upright.
\end{itemize}

Tasks I, IV, V, and VI are scored binary, and are the four tasks pooled for the statistical analysis in Section~\ref{sec:main_results}. Tasks II and III use the additive rubrics above.

\subsection{Baseline Implementation Details}
\label{app:strong_baseline_details}

\begin{table*}[t!]
\centering
\caption{\textbf{Baseline configurations.}
GE denotes Genie Envisioner and RDP denotes Reactive Diffusion Policy.
Camera access is identical across methods within
each task (Section~\ref{sec:experiments}).}
\label{tab:supp_strong_baseline_setup}
\vspace{-0.2cm}
\footnotesize
\setlength{\tabcolsep}{5pt}
\renewcommand{\arraystretch}{1.2}

\begin{tabularx}{\linewidth}{
    @{}l
    >{\raggedright\arraybackslash}X
    >{\raggedright\arraybackslash}X
    >{\raggedright\arraybackslash}X@{}
}
\toprule
Model & Tactile input & World model & Policy type \\
\midrule

$\pi_{0.5}$
& None
& None
& Vision-language-action policy \\

ViTacFormer
& 6D fingertip wrenches
& None (auxiliary tactile forecast)
& ACT/CVAE policy \\

RDP
& 6D fingertip wrenches
& None
& Tactile diffusion policy \\

GE
& None
& Vision
& Video-action model \\

\midrule
\rowcolor[HTML]{EDF3FD}
\dextacwam{}
& Dense per-finger tactile maps
& Visuo-tactile
& World-action model \\

\bottomrule
\end{tabularx}
\end{table*}

We summarize the baseline configurations in Table~\ref{tab:supp_strong_baseline_setup} and their parameter counts in Table~\ref{tab:model_size}. The four baselines are chosen to span the design space around \methodname{}: a generalist VLA policy, a tactile policy with an auxiliary tactile forecasting head, a tactile diffusion policy, and a vision-only WAM.

\paragraph{$\pi_{0.5}$.}
$\pi_{0.5}$ is a generalist robot policy that represents large-scale robot policy pretraining. It receives no tactile input, and therefore isolates how far a strong pretrained visuomotor prior goes on contact-rich dexterous tasks.

\paragraph{ViTacFormer.}
ViTacFormer is an ACT/CVAE policy that consumes vision together with the ten fingertip 6-D force/torque readings ($[18,120]$) and autoregressively forecasts future tactile tokens through an auxiliary head inside the policy. For deployment on our 22-DoF platform, we retain the released inputs and architecture apart from resizing the action head from 17 to 22 DoF, and align temporal aggregation with the 30\,Hz training timeline. The resulting scores are reported in Table~\ref{tab:success_rate}.

\paragraph{Reactive Diffusion Policy.}
RDP uses head-view visual observations together with the ten fingertip 6-D wrenches as tactile observations, and predicts actions with a diffusion policy. It is the closest baseline to \methodname{} in terms of raw sensor access: it receives the same tactile signal, but consumes it as a downstream policy input rather than modeling contact evolution as part of the predicted world state. It is therefore the reference point for the ablation in Section~\ref{sec:ablations}.

\paragraph{Genie Envisioner.}
GE is a vision-based WAM that predicts actions through a video-action modeling framework without tactile input. It tests whether visual world modeling alone suffices when contact state and force are only partially observable from cameras. GE and \methodname{} share the same backbone scale and action-expert size (Table~\ref{tab:model_size}), so the comparison between them isolates the tactile pathway rather than model capacity.

\paragraph{Action expert capacity.}
\methodname{} consists of a 2B world-model backbone and a 160M action expert. Unlike $\pi_{0.5}$, whose 400M action expert benefits from large-scale action pretraining, our action expert is initialized from scratch and sees only the roughly 100 task-specific episodes. Pretraining the action expert is a clear direction for further gains and remains future work.

\subsection{Held-Out Bowl Configurations}
\label{app:generalization_rollouts}
\begin{figure*}[p]
    \centering
    \includegraphics[width=\linewidth]{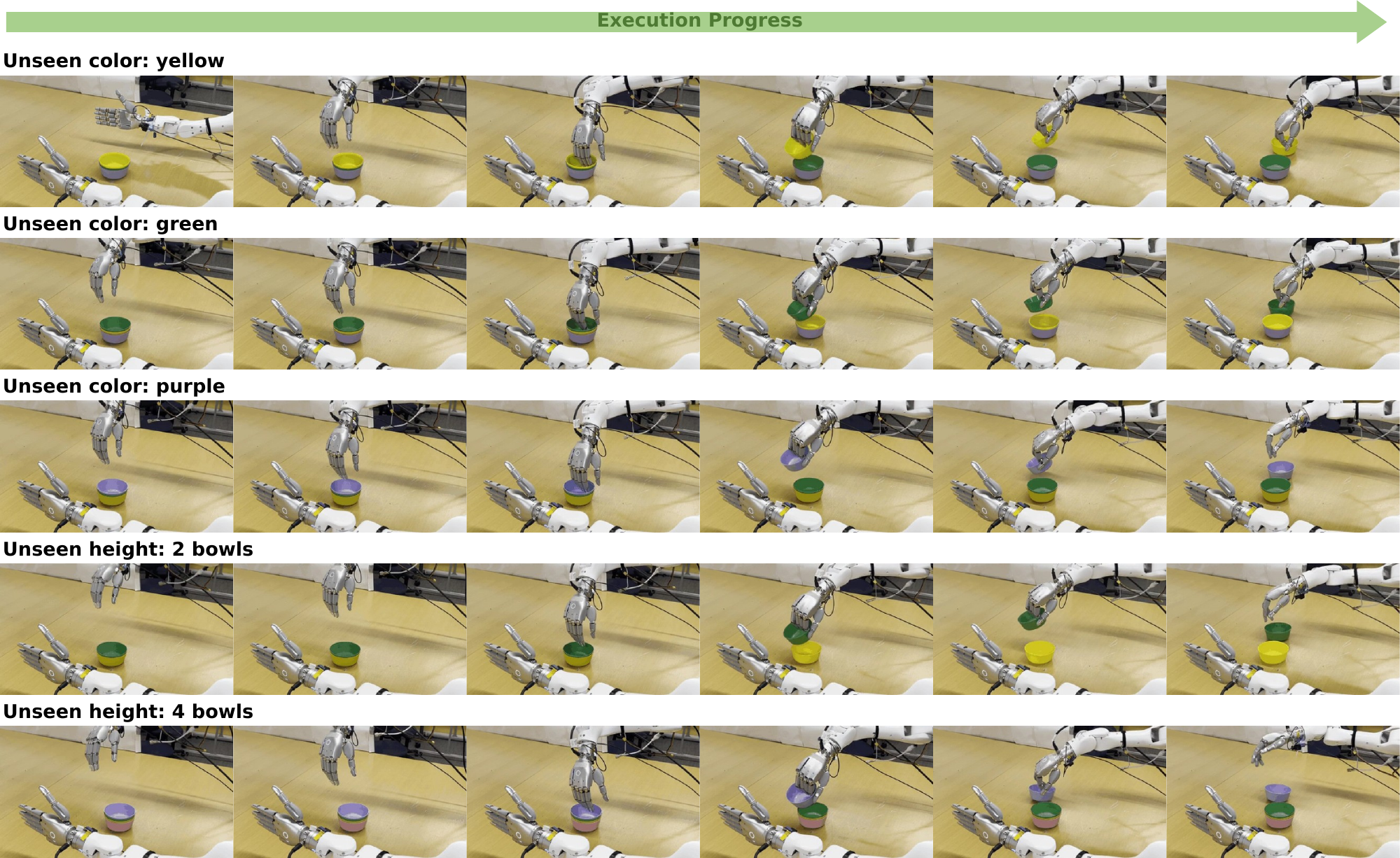}
    \vspace{-0.5cm}
    \caption{\textbf{Rollouts on held-out Bowl configurations.}
    Keyframes progress from left to right for three unseen bowl colors and two unseen stack heights, none seen during training, and are sampled from autonomous closed-loop executions.
    In all cases, the policy reaches the rim, establishes thumb--index contact, and separates the top bowl.}
    \label{fig:generalization_rollouts}
\end{figure*}

Figure~\ref{fig:generalization_rollouts} shows representative rollouts for the held-out Bowl configurations evaluated in Section~\ref{sec:main_results}.
The policy is trained only on three-bowl stacks in pink, blue, and orange.
At test time, we vary either bowl color or stack height, changing the visual appearance, rim height, and resulting grasp configuration.

\subsection{Visual Prediction Retention Protocol}
\label{app:vision_retention_protocol}

For each task in Table~\ref{tab:vision_retention}, the visuo-tactile world model and its vision-only counterpart use the same training episodes, held-out episodes, camera views, and number of finetuning steps; the only difference is whether tactile views are enabled. Both models are evaluated on future-RGB prediction. Because episode difficulty varies by roughly $2$\,dB, we compute tactile-minus-vision-only differences for matched episodes rather than comparing unpaired means. Each held-out episode is one statistical observation, with its three camera views averaged because views from the same episode are not independent.

\begin{table}[t!]
\centering
\caption{\textbf{Model size comparison.}
We report the approximate number of parameters in the backbone/world model, action expert, and full model. }
\label{tab:model_size}
\vspace{-0.2cm}
\footnotesize
\setlength{\tabcolsep}{4pt}
\renewcommand{\arraystretch}{1.15}

\begin{tabularx}{\linewidth}{
    @{}>{\raggedright\arraybackslash}X rrr@{}
}
\toprule
Model
& \shortstack{Backbone /\\world model}
& \shortstack{Action\\expert}
& Total \\
\midrule

$\pi_{0.5}$
& 2.5B & 400M & 2.9B \\

ViTacFormer
& -- & -- & 100M \\

RDP
& -- & -- & 200M \\

Genie Envisioner
& 2B & 160M & 2.16B \\

\midrule
\rowcolor[HTML]{EDF3FD}
\dextacwam{}
& 2B & 160M & 2.16B\\

\bottomrule
\end{tabularx}
\end{table}

\subsection{Multi-Finger Tactile Encoder Details}
\label{app:tactile_encoder_impl}

Our multi-finger tactile encoder consists of a grayscale-to-RGB projection, a reused visual VAE that maps each per-finger tactile map into a spatial latent grid, and a finger- and pose-aware tactile compressor that maps five per-finger latents into one compact per-hand tactile latent. This design reuses the spatial prior of a pretrained video model while preserving multi-finger contact structure.

\begin{figure*}[t]
    \centering
    \includegraphics[width=\linewidth]{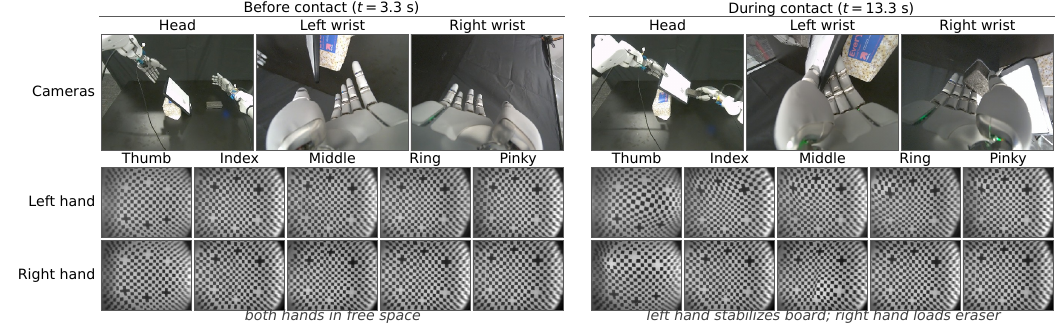}
    \vspace{-0.5cm}
    \caption{\textbf{Raw visuo-tactile observations before and during contact on the two-hand wipe task.}
    Each block shows one instant from a teleoperated episode, including the head and two wrist-camera views together with all ten fingertip tactile observations.
    Before contact, both hands are in free space and the marker grids remain largely undeformed.
    During wiping, the left hand stabilizes the board while the right hand loads the eraser, producing localized compression and shear on the engaged fingertips.
    All tactile panels share the same grayscale range.}
    \label{fig:raw_tactile_observations}
\end{figure*}

Figure~\ref{fig:raw_tactile_observations} shows what these observations look like before the encoder. Each fingertip returns a grayscale image of a deformable marker grid, and contact appears as a localized compression and shear of that grid rather than as a scalar force reading. This spatial image structure makes a pretrained visual VAE a natural starting point for tactile encoding.

\paragraph{Reused visual VAE encoder.}
Each tactile observation is represented as a single-channel dense tactile map for each finger. Since the pretrained LTX visual VAE expects three-channel inputs, we first apply a lightweight $1{\times}1$ grayscale-to-RGB adapter:
\begin{equation}
    x_{t,i}^{rgb}
    =
    \phi_{g\rightarrow rgb}(o_{t,i}^{\tau}),
\end{equation}
where $\phi_{g\rightarrow rgb}$ is initialized so that its output is equivalent to repeating the grayscale tactile map across three channels. The resulting pseudo-RGB tactile map is then passed into the frozen visual VAE encoder:
\begin{equation}
    z_{t,i}^{\tau}
    =
    E_{\mathrm{vae}}\!\left(
    \phi_{g\rightarrow rgb}(o_{t,i}^{\tau})
    \right).
\end{equation}
The VAE encoder is kept frozen, so tactile adaptation happens through the grayscale-to-RGB adapter and the downstream tactile compressor rather than by changing the pretrained VAE latent space.

\paragraph{Full tactile compressor.}
For each hand, the VAE produces five per-finger tactile latent grids. The full compressor aggregates these five finger latents into one per-hand latent grid:
\begin{equation}
    \hat{z}^{\tau,\cdot}
    =
    A_\theta(Z^{\tau,\cdot}, q),
    \qquad
    \hat{z}^{\tau,\cdot}
    \in
    \mathbb{R}^{C \times T_{\mathrm{lat}} \times H' \times W'} ,
\end{equation}
where $\cdot \in \{L,R\}$ denotes the left or right hand. In our implementation, $C=128$ and the spatial latent grid is $H' \times W' = 6 \times 8$. The temporal length is aligned with the visual latent track:
\begin{equation}
    T_{\mathrm{lat}}
    =
    \mathrm{mem\_size}
    +
    \left\lfloor
    \frac{\mathrm{chunk}}{8}
    \right\rfloor
    +
    1 .
\end{equation}

The compressor performs per-cell set encoding. For each spatio-temporal cell $(t,h,w)$, we form a six-token set containing one hand query and five finger tokens:
\begin{equation}
    \mathcal{S}_{t,h,w}
    =
    \{q_{\mathrm{hand}}, z_{t,1,h,w}^{\tau}, \ldots, z_{t,5,h,w}^{\tau}\}.
\end{equation}
Each finger token receives a learnable finger identity embedding $e_i$, and the hand query receives a learnable spatial bias $p_{h,w}$. A small self-attention set encoder is applied to this six-token set, and retaining its output hand-query token compresses the five finger tokens into one per-cell hand representation:
\begin{equation}
\scalebox{0.8}{$
    \tilde{z}_{t,h,w}^{\mathrm{set}}
    =
    \mathrm{SetEnc}
    \left(
    q_{\mathrm{hand}} + p_{h,w},
    z_{t,1,h,w}^{\tau}+e_1,
    \ldots,
    z_{t,5,h,w}^{\tau}+e_5
    \right).
    $}
\end{equation}
To stabilize training, the set-encoder output is added as a residual correction to a softmax-weighted finger average:
\begin{equation}
    \tilde{z}_{t,h,w}
    =
    \sum_{i=1}^{5}
    \mathrm{softmax}(\ell)_i
    z_{t,i,h,w}^{\tau}
    +
    \alpha
    \tilde{z}_{t,h,w}^{\mathrm{set}},
\end{equation}
where the finger logits $\ell$ and residual gain $\alpha$ are initialized so that the compressor starts from a uniform finger average.

The full compressor also injects hand pose into the hand query. Given hand pose $q_t$, a pose encoder maps it into the tactile latent dimension:
\begin{equation}
    q_{\mathrm{hand}}^{t,h,w}
    \leftarrow
    q_{\mathrm{hand}}
    +
    \alpha_q E_q(q_t)
    +
    p_{h,w},
\end{equation}
where $\alpha_q$ is a learnable scalar initialized to zero. Finally, the fused hand-level tactile grid is refined by lightweight divided space-time attention, with temporal attention over $T_{\mathrm{lat}}$ and spatial attention over $H'W'$. The refinement output is added through another zero-initialized residual gate:
\begin{equation}
    \hat{z}^{\tau}
    =
    \tilde{z}^{\tau}
    +
    \alpha_t
    \left(
    \tilde{z}^{\tau,\mathrm{refined}}
    -
    \tilde{z}^{\tau}
    \right).
\end{equation}
This refinement helps model contact transitions such as slip, regrasping, handover, and sustained wiping contact.

\paragraph{Cross-attention pooling baseline.}
For the compressor recall comparison, we use a cross-attention pooling baseline that keeps the same frozen visual VAE encoder and the same per-finger tactile inputs, but replaces the full compressor with a single hand-query cross-attention layer. For each spatio-temporal cell, the baseline uses one hand query and five finger tokens:
\begin{equation}
    q = q_{\mathrm{hand}} + p_{h,w},
    \qquad
    k_i = v_i = z_{t,i,h,w}^{\tau} + e_i.
\end{equation}
The fused output is produced by one multi-head attention pooling operation:
\begin{equation}
    z_{t,h,w}^{\mathrm{pool}}
    =
    \mathrm{MHA}(q, \{k_i\}_{i=1}^{5}, \{v_i\}_{i=1}^{5}).
\end{equation}
As in the full compressor, the attention output is added as a residual correction to a softmax-weighted finger average:
\begin{equation}
    \hat{z}_{t,h,w}^{\tau}
    =
    \sum_{i=1}^{5}
    \mathrm{softmax}(\ell)_i
    z_{t,i,h,w}^{\tau}
    +
    \alpha
    z_{t,h,w}^{\mathrm{pool}}.
\end{equation}
This baseline does not use pose injection or divided space-time refinement, and therefore tests whether simple per-cell attention pooling is sufficient for preserving multi-finger contact information during five-finger-to-one-hand compression.

\paragraph{Intermediate ablation variants.}
The component ablation in Table~\ref{tab:adapter_recall} interpolates between cross-attention pooling and the full compressor. ``Self-attention pooling'' applies the six-token set encoder above but omits pose injection and space-time refinement. ``Self-attention $+$ pose injection'' adds the pose-conditioned hand query, whereas ``Self-attention $+$ temporal refinement'' instead adds divided space-time attention; the full compressor uses both. The ``w/o finger-identity embeddings'' variant is the full compressor with the learnable per-finger embeddings $e_i$ removed, so finger tokens become exchangeable and the set encoder can no longer attribute contact to a specific finger. All variants share the same frozen VAE encoder and per-finger inputs.
\subsection{Training Details}
\label{app:training_details}

We summarize \methodname{}'s training hyperparameters in Table~\ref{tab:training_details}. The training pipeline has three stages: Stage 1 learns a reusable tactile encoder while keeping its visual VAE frozen, Stage 2 directly finetunes the pretrained video model for cross-modal world modeling on each downstream task, and Stage 3 trains a freshly initialized action expert on the same task demonstrations.\looseness-1

\begin{table*}[t!]
\centering
\caption{\textbf{Training configurations across the three stages.}
Stage 1 adapts the tactile encoder; Stages 2 and 3 are trained per task.}
\label{tab:training_details}
\vspace{-0.2cm}
\footnotesize
\setlength{\tabcolsep}{5pt}
\renewcommand{\arraystretch}{1.15}

\begin{tabularx}{0.95\linewidth}{
    @{}>{\raggedright\arraybackslash}p{0.22\linewidth}
    *{3}{>{\raggedright\arraybackslash}X}@{}
}
\toprule
\rowcolor[HTML]{EDF3FD}
Setting
& \textbf{Stage 1}\newline Tactile-encoder adaptation
& \textbf{Stage 2}\newline World-model fine-tuning
& \textbf{Stage 3}\newline Action fine-tuning \\
\midrule

\rowcolor{gray!20} \multicolumn{4}{@{}l}{\textit{Model and data}} \\
\addlinespace[2pt]

Trainable modules
& GrayToRGB, tactile compressor, pose encoder, TimeSformer,
  auxiliary heads
& DiT world-model backbone
& DiT backbone and action expert \\

Frozen modules
& LTX visual VAE
& Tactile encoder, LTX VAE, text encoder
& Tactile encoder, LTX VAE, text encoder \\

Initialization
& Tactile compressor warm start
& Pretrained video DiT and Stage-1 tactile encoder
& Per-task Stage-2 checkpoint and new action expert \\

Data
& Diverse bimanual manipulation corpus
& Task demonstrations
& Task demonstrations \\

\midrule
\rowcolor{gray!20} \multicolumn{4}{@{}l}{\textit{Optimization and execution}} \\
\addlinespace[2pt]

Batch size per process
& 16 & 16 & 16 \\

Effective batch size
& 16 & 64 & 64 \\

Distributed strategy
& Single GPU
& DDP
& DeepSpeed ZeRO-2 or single GPU \\

Optimizer
& AdamW & AdamW & AdamW \\

AdamW $(\beta_1,\beta_2)$
& $(0.9, 0.95)$
& $(0.9, 0.95)$
& $(0.9, 0.95)$ \\

Weight decay
& $1.0\times10^{-4}$
& $1.0\times10^{-5}$
& $1.0\times10^{-5}$ \\

Learning rate
& $1.0\times10^{-4}$
& $3.0\times10^{-4}$
& $5.0\times10^{-5}$ \\

Learning-rate schedule
& Constant; 1K-step warmup
& Constant; 1K-step warmup
& Constant; 1K-step warmup \\

Gradient clipping
& -- & 1.0 & 1.0 \\

Gradient checkpointing
& Disabled & Enabled & Enabled \\

Precision
& BF16; TF32 enabled
& BF16
& BF16 \\

Checkpoint saving
& Every 5K steps
& Every 5K steps
& Every 10K steps \\

\midrule
\rowcolor{gray!20} \multicolumn{4}{@{}l}{\textit{Sampling and objectives}} \\
\addlinespace[2pt]

Sampling protocol
& $T_{\mathrm{choices}}\in\{1,9\}$
& Chunk: 9\newline Action chunk: 54\newline Memory frames: 4
& Chunk: 9\newline Action chunk: 54\newline Memory frames: 4 \\

Caption dropout
& -- & 0.06 & 0.06 \\

Conditioning-frame noise
& --
& 0.1 during training
& 0.1 during training;\newline clean memory at inference \\

Loss weights
& $\lambda_{\mathrm{loc}}=1.0$;\newline
  $\lambda_{\mathrm{loc\_pre}}=0.3$
& $\lambda_v=1.0$;\newline $\lambda_\tau=1.0$
& Action loss scale: 1.0 \\

Flow weighting
& Square-root-magnitude-aware
& None
& None \\

\bottomrule
\end{tabularx}
\end{table*}

Stage 1 freezes the LTX visual VAE and trains the tactile representation modules with a flow reconstruction objective. Stage 2 freezes the tactile encoder and finetunes the DiT world-model body with the visuo-tactile co-denoising objective. Stage 3 warm-starts from the corresponding Stage-2 world-model checkpoint, initializes a fresh action expert, and trains the action model with the unified 194-dimensional flow-matching action objective.
\subsection{Ablation Variant Implementation Details}
\label{app:ablation_impl}

We describe how each variant for ablation results reported in Section~\ref{sec:ablations} is constructed.

\paragraph{Direct tactile conditioning (w/o tactile world modeling).}
This variant keeps the tactile encoder, tactile observations, action expert, action space, and training data of \methodname{}, and removes only tactile prediction from the world model. The action expert is conditioned on encoded tactile features directly instead of on predicted tactile latents. It therefore measures the contribution of predictive tactile world modeling while holding tactile information content fixed.\looseness-1

\paragraph{W/o per-modality K/V RMS normalization.}
This variant removes the parameter-free RMS normalization applied to the multi-view visual and tactile keys/values before the action cross-attention. Everything else is unchanged. The action expert fails to converge: open-loop action predictions do not track ground truth, so we do not proceed to closed-loop evaluation.

\paragraph{Reference configuration.}
All variants above are defined by subtraction from the same reference: the full \methodname{} model, which uses a visuo-tactile world model and conditions the action model on predictive tactile latents. Two properties of that reference are worth stating explicitly, since they are easily conflated. At deployment, the action model takes \emph{no} force input. Force enters only during training, as the per-finger force block of the unified action target described in Section~\ref{sec:method}. Tactile sensing is therefore separated from force-conditioned control: the policy acts on predictive tactile representations, while the auxiliary force supervision keeps the action model from ignoring touch.

\subsection{Failure Case Analysis}
\label{appendix:failure_analysis}

We further analyze representative failure modes observed during real-world evaluation, broken down by task.

\paragraph{Cube Place w/ Occlusion.}
In this task, the wrist cameras are disabled and the robot arm often occludes the cube from the head camera, making precise visual localization difficult. As a result, policies without tactile feedback frequently fail to localize the cube accurately after the hand approaches the object. In particular, $\pi_{0.5}$ often fails at the grasping stage because visual observations alone provide insufficient information once the cube is partially or fully occluded by the arm. In contrast, \methodname{} can use fingertip tactile observations to detect and correct contact with the cube, which improves robustness during the final stage of grasp acquisition. This failure mode highlights the value of tactile feedback when visual information is degraded by self-occlusion.

\paragraph{Cube Handover.}
Cube Handover serves as a standard dexterous manipulation task and provides a contrast to the more contact-rich tasks in our evaluation. The dominant failure mode is sensitivity to the cube's initial pose. When the initial pose falls outside the training distribution, the left hand may grasp the cube at an inaccurate location or orientation, and this error propagates to the subsequent handover. A promising direction is tactile-informed reinforcement learning for closed-loop correction: once contact is established, fingertip observations provide more direct and detailed information than wrist-camera images about the cube's pose relative to the fingers and the stability of the grasp, enabling the policy to adjust the grasp before transfer. Even with an action expert trained from scratch, only four hours of tactile-representation pretraining, and no reinforcement-learning post-training, \methodname{} achieves the highest score among all evaluated baselines on this task, highlighting the benefit of predictive tactile world modeling.

\paragraph{Two-Hand Wipe.}
The two-hand wipe task requires both stable physical interaction and high-level task completion. \methodname{} benefits from tactile feedback when maintaining contact with the whiteboard and manipulating the eraser, but its failures are often associated with incomplete task execution. In particular, \methodname{} may wipe part of the black markings but fail to fully clean the board, suggesting limitations in instruction following and long-horizon task progress monitoring compared with large-scale pretrained VLA models. In contrast, $\pi_{0.5}$ often fails because it does not maintain a stable hold on the whiteboard with the left hand. Without tactile feedback, the policy may not detect that the board is slipping or that the left hand is applying insufficient stabilizing force. These failures typically cause the whiteboard to move away during wiping, preventing successful task completion. Thus, this task exposes complementary limitations: \methodname{} is more robust to contact instability, while $\pi_{0.5}$ benefits from broader pretrained visuomotor priors but struggles with sustained tactile contact.

\paragraph{Tongs.}
This task requires precise control of force transmitted through a tool: the robot must hold and close the tongs firmly enough to lift the cherry tomato without allowing it to slip or crushing it. The baselines commonly fail to grasp the tongs stably or to apply an appropriate closing force around the tomato. In contrast, \methodname{} consistently closes the tongs around the tomato, transfers it to the plate, releases it, and returns the tongs to the table. This behavior highlights the benefit of predicting contact dynamics when the relevant grasp state is mediated by a tool and is difficult to infer visually.

\paragraph{Bowl Unstacking.}
\methodname{} coordinates the right thumb, index finger, and middle finger to establish stable opposing contacts and separate the top bowl from the stack. Baseline failures arise when this multi-finger contact is unstable or when spatial inaccuracies cause the fingers to miss or poorly align with the bowl rim. The task therefore requires both accurate spatial alignment and sufficient finger-specific friction to lift one bowl without disturbing the remaining stack.

\paragraph{Bottle Cap Unscrewing.}
Unscrewing requires the left hand to stabilize the bottle while the right hand, particularly the thumb, maintains friction with the cap and applies torque in a consistent direction. Baselines often grasp the bottle too weakly with the left hand, causing it to tilt, or apply unstable right-hand force directions that dislodge or drop the bottle. \methodname{} more consistently holds the bottle upright with the left hand while using the right thumb and fingers to rotate and remove the cap, demonstrating coordinated control of stabilizing force, friction, and rotational contact.

\subsection{Success-Only Training and Recovery}
\label{app:success_bias}

Predictive tactile modeling substantially improves contact-rich manipulation by enabling the policy to better exploit contact information during nominal execution. Closing the remaining gap toward more reliable behavior likely requires a second capability: self-correction after mistakes. Our success-only training data provide little supervision for such recovery behavior, leaving failure states such as missed grasps, slip, and dropped objects out of distribution for both the world model and the policy.\looseness-1

This limitation is shared by imitation-based manipulation systems more broadly: once execution deviates from the demonstration distribution, compounding errors and the lack of recovery experience can substantially degrade closed-loop behavior~\cite{ross2011dagger}. Our qualitative inspection suggests a similar pattern: visuo-tactile predictions generally remain plausible before the first execution error, while both prediction and action quality deteriorate after the policy enters unseen failure states. Incorporating failure and recovery experience~\cite{peng2026fact} through perturbation, intervention, or reinforcement learning is a promising direction for addressing this limitation.

\subsection{World-Model Prediction Visualization}
\label{app:wm_prediction_visualization}

Figure~\ref{fig:wm_pred_visual_tactile} in the main paper shows joint visual and tactile predictions for one fingertip.
Here, we expand the same rollout to all ten fingers and all tactile-flow channels, focusing on two representative instants: $t=24.2$\,s, near peak wiping shear, and $t=33.4$\,s, later in the stroke.

\begin{figure*}[t]
    \centering
    \includegraphics[width=0.99\linewidth]{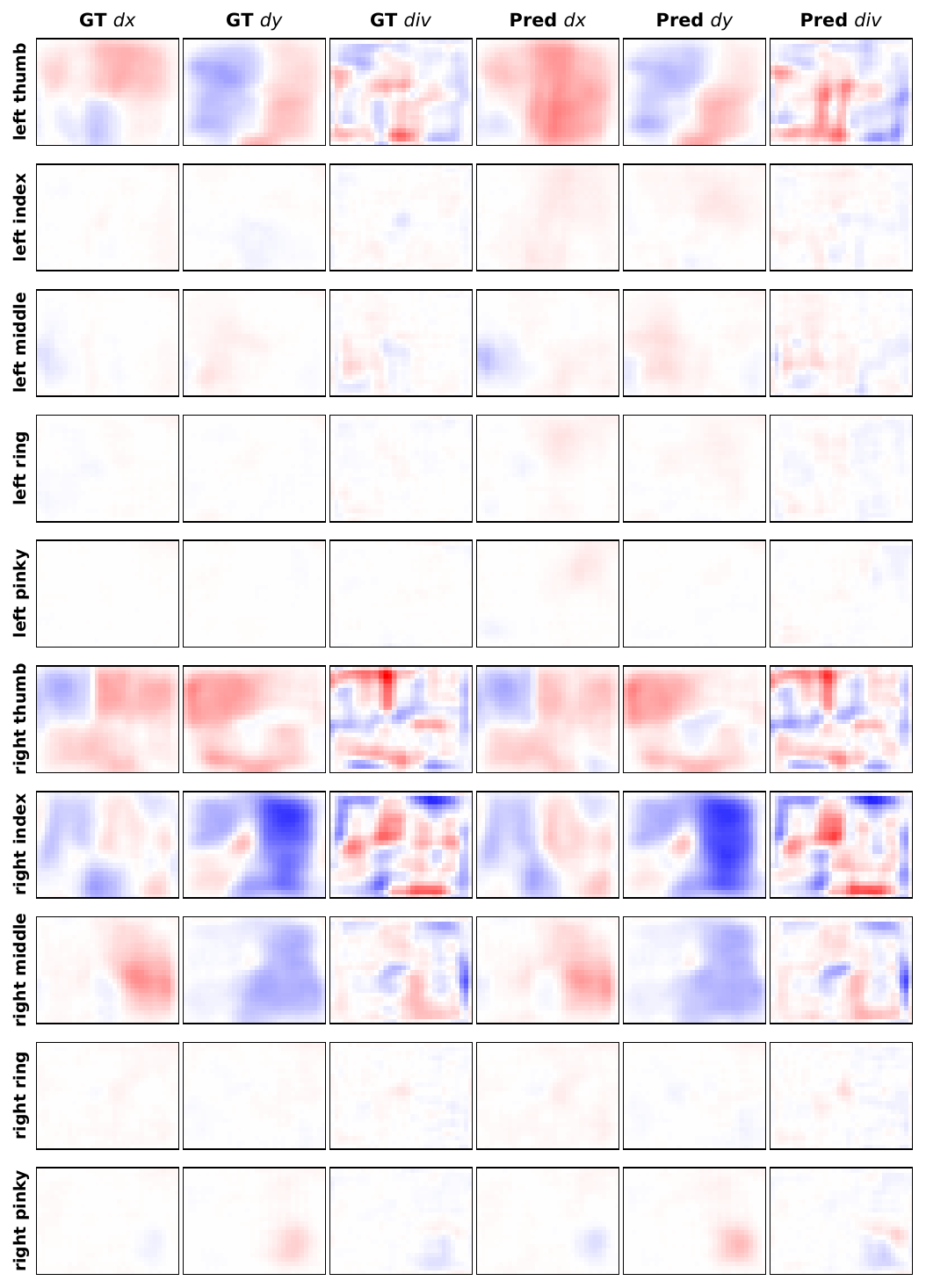}
    \vspace{-0.3cm}
    \caption{\textbf{Per-finger tactile-flow prediction near peak wiping shear.}
    Ground-truth and predicted $dx$, $dy$, and divergence fields are shown for all ten fingers at $t=24.2$\,s.
    All panels use the same symmetric color scale determined from the ground truth, so near-zero intensity indicates weak or absent contact rather than independent rescaling.}
    \label{fig:tactile_flow_per_finger_24s}
\end{figure*}

\begin{figure*}[t]
    \centering
    \includegraphics[width=0.99\linewidth]{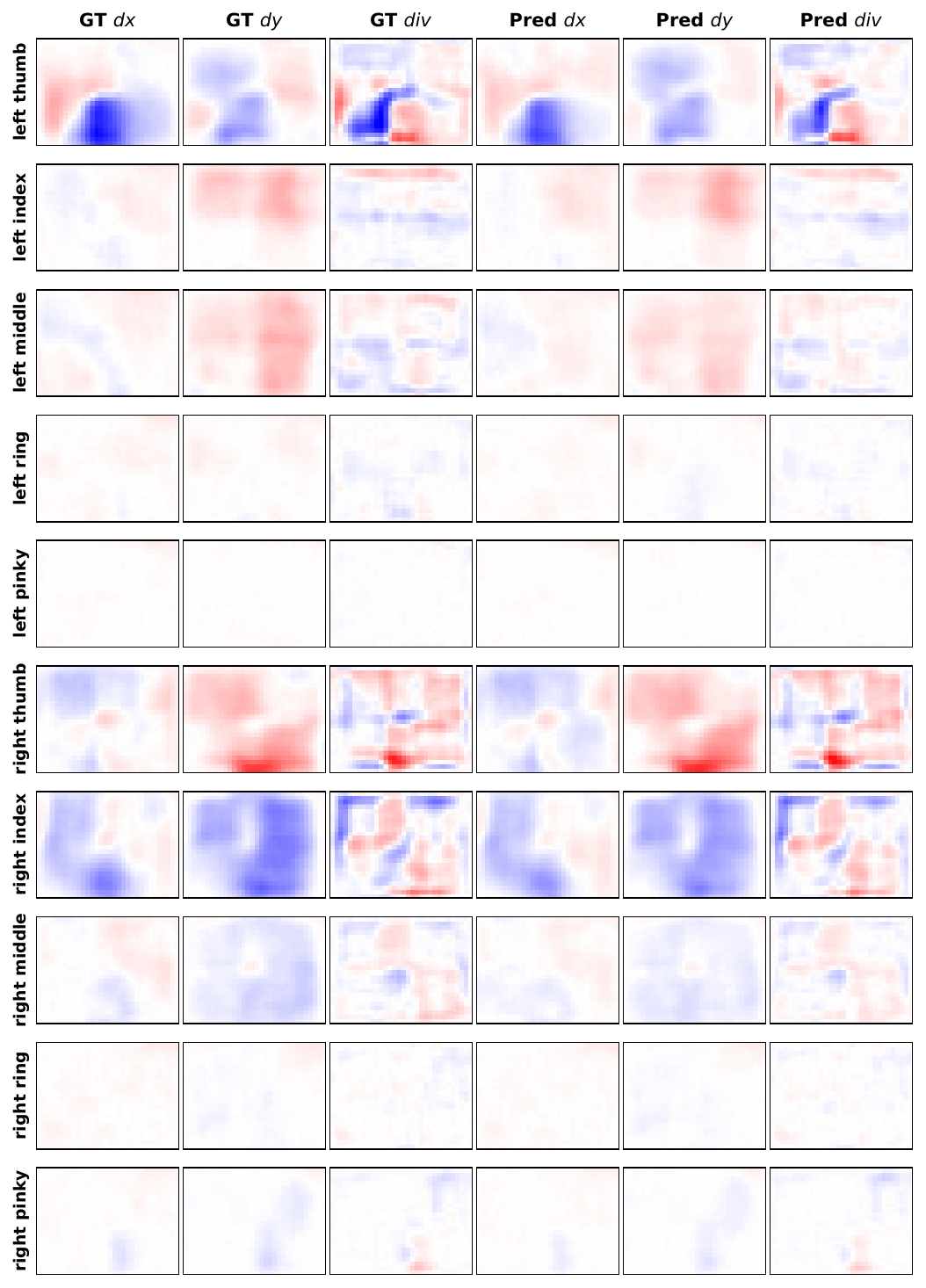}
   \vspace{-0.3cm}
    \caption{\textbf{Per-finger tactile-flow prediction later in the wiping stroke.}
    Ground-truth and predicted tactile-flow fields are shown for all ten fingers at $t\!=\!33.4$\,s.
    The dominant contact pattern and the change in shear direction are preserved as interaction evolves\looseness-1.}
    \label{fig:tactile_flow_per_finger_33s}
\end{figure*}

Figures~\ref{fig:tactile_flow_per_finger_24s} and~\ref{fig:tactile_flow_per_finger_33s} show the per-finger tactile-flow predictions.
The model captures the dominant deformation patterns on the fingers carrying contact, while fingers with little or no contact remain near zero in both prediction and ground truth.
The change in shear direction across the two instants further indicates that the model captures evolving contact dynamics rather than simply copying a static tactile imprint. For example, the left-thumb $dx$ field changes sign between the two instants, and the prediction follows the same transition.

\subsection{Why Continual Vision-to-Touch Learning Works}
\label{app:tactile_latent_tsne}
To understand how continual vision-to-touch learning acquires useful tactile representations with limited data, we analyze the features produced by the reused visual VAE and the tactile compressor. We examine task discrimination before and after compression, whether within-task structure reflects interaction progress, and the relationship between visual and tactile latent spaces.

\paragraph{Evaluation protocol.}
We sample temporal tactile clips from held-out episodes of all six tasks using seed $42$, obtaining $1{,}350$ right-hand examples across multiple episodes and temporal positions ($150$ per bimanual task and $300$ per single-hand task). For each clip, we extract two representations: pooled $128$-D right-thumb features from the learned grayscale-to-RGB projection and frozen visual VAE, and pooled $128$-D hand features from the compressor after fusion of all five fingertips. We visualize each representation independently using PCA to $50$ dimensions followed by t-SNE with perplexity $30$ and seed $42$. These plots illustrate structure within each representation. All quantitative metrics are computed in the original $128$-D feature space; $5$-nearest-neighbour and linear-probe classification use stratified five-fold evaluation.\looseness-1

\begin{figure*}[t!]
    \centering
    \includegraphics[width=0.95\linewidth]{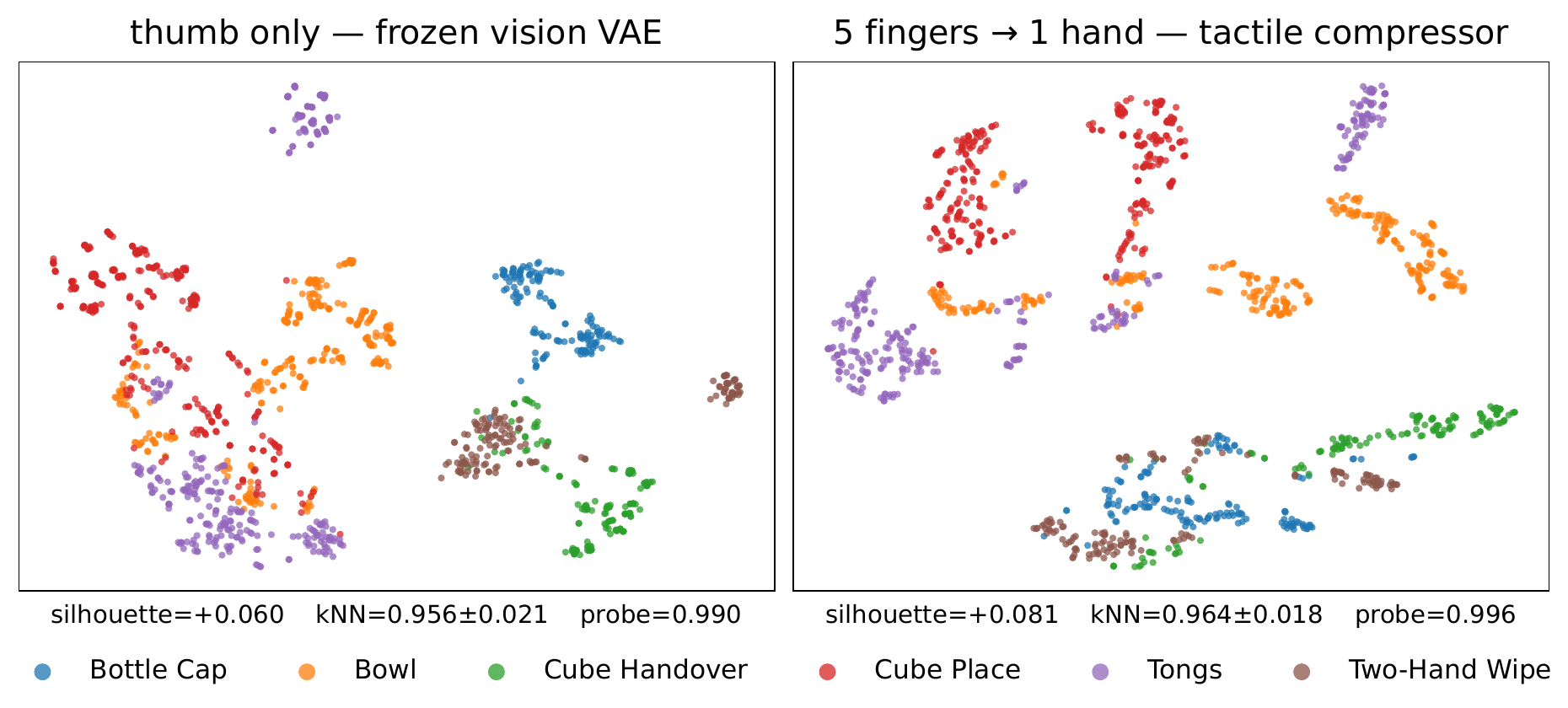}
    \vspace{-0.3cm}
    \caption{\textbf{The tactile compressor preserves task-relevant information.}
    Each point is one tactile clip and colours denote the six tasks.
    Left: the frozen visual-VAE representation of the right thumb.
    Right: the compressor output after mapping all five fingertips into one hand latent.
    Task separation remains after compression.
    The plots visualize within-representation task structure; metrics are computed in the original $128$-D space.}
    \label{fig:tactile_tsne_thumb_vs_adapter}
\end{figure*}

\paragraph{Can a frozen visual VAE provide useful tactile representations?}
As shown in Figure~\ref{fig:tactile_tsne_thumb_vs_adapter}, the right-thumb features distinguish the six tasks, achieving $95.6\%$ $5$-nearest-neighbour accuracy and $99.0\%$ linear-probe accuracy, compared with a $16.7\%$ uniform chance level and a $22.2\%$ majority-class baseline. These results show that a frozen pretrained visual VAE, combined with a learned grayscale-to-RGB projection, can provide task-discriminative tactile features without training a tactile VAE from scratch.

\paragraph{Does $5{:}1$ compression preserve the structure needed for tactile world modeling?}
The compressed hand representation achieves $96.4\%$ $5$-nearest-neighbour accuracy and $99.6\%$ linear-probe accuracy, with a silhouette score of $0.081$, compared with $0.060$ for the right-thumb representation. Thus, the hand latents remain strongly task-discriminative after $5{:}1$ compression, making tactile world modeling efficient while preserving the task-relevant structure.  

\paragraph{Do tactile latents reflect interaction progress?}
Task discrimination alone does not establish whether tactile features capture changes within an interaction. We therefore analyze each task's right-thumb features separately and visualize them by normalized clip phase. In the original $128$-D space, we fit $k$-means with $k\in{2,3,4}$ and select $k$ using the silhouette score. We measure the association between cluster membership and clip phase using ANOVA $R^2$, and agreement with episode identity using adjusted mutual information (AMI).

\begin{figure*}[t!]
    \centering
    \includegraphics[width=0.95\linewidth]{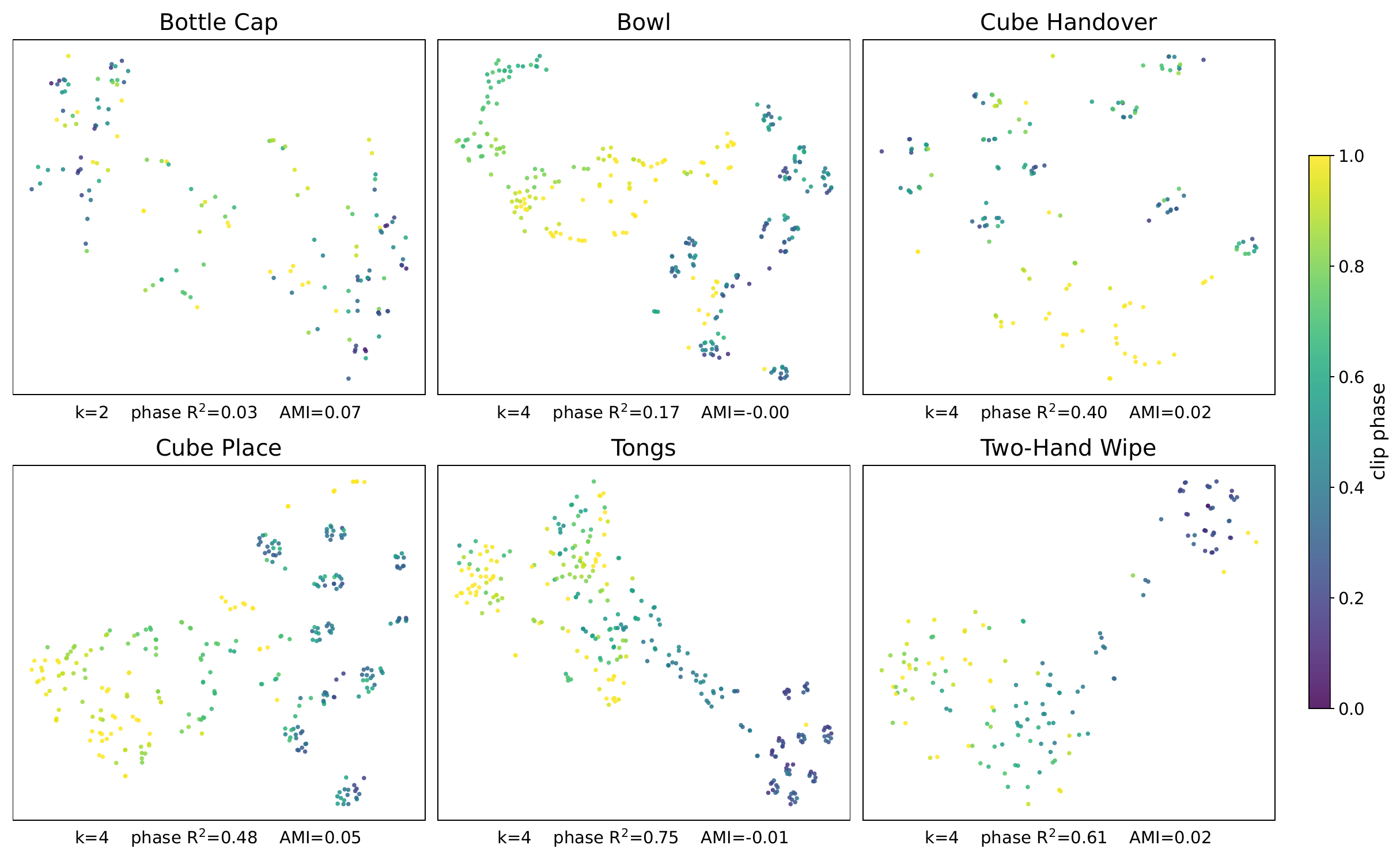}
        \vspace{-0.3cm}
\caption{\textbf{Tactile latent structure tracks interaction progress.}
    Each panel shows right-thumb tactile latents from one task; colour indicates when each clip occurs from the start to the end of an episode.
    In five of six tasks, clusters align with interaction progress but not with episode identity, indicating within-task temporal structure rather than recording-specific artifacts.
    Bottle Cap is the exception because unscrewing is cyclic and cannot be described by a single start-to-end phase.}
    \label{fig:tactile_tsne_contact_phase}
\end{figure*}

As shown in Figure~\ref{fig:tactile_tsne_contact_phase}, cluster membership is associated with interaction phase to varying degrees: $R^2$ is $0.75$ for Tongs, $0.61$ for Two-Hand Wipe, $0.48$ for Cube Place, $0.40$ for Cube Handover, and $0.17$ for Bowl. Agreement with episode identity is low across all tasks ($\mathrm{AMI}\leq0.07$). These results suggest that within-task latent structure captures aspects of interaction progress beyond recording identity. Bottle Cap shows little association with normalized phase ($R^2=0.03$) because cyclic unscrewing is not described well by a single start-to-end phase.

\paragraph{Relationship between visual and tactile latent spaces.}
Visual and tactile latents remain well separated, with a between- to within-modality distance ratio of approximately $89$. The observed benefits of reusing the visual encoder therefore do not imply that visual and tactile observations map to a shared, modality-invariant representation. A pretrained visual encoder can provide useful tactile features while the two modalities retain distinct feature distributions.

\end{document}